\documentclass[runningheads]{llncs}
\usepackage{fontawesome5}

\usepackage{eccv}

\usepackage{eccvabbrv}

\usepackage{graphicx}
\usepackage{booktabs}
\usepackage{tabularx}

\usepackage[accsupp]{axessibility}  

\usepackage{hyperref}

\usepackage{orcidlink}

\newcommand\modelname{TIP}
\usepackage[ruled,linesnumbered]{algorithm2e}
\usepackage{array}
\newcolumntype{P}[1]{>{\centering\arraybackslash}p{#1}}

\usepackage{caption}
\usepackage{footnote}

\newcommand{\customfootnote}[1]{
    \begingroup
    \renewcommand{\thefootnote}{}
    \footnotetext{#1}
    \addtocounter{footnote}{-1}
    \endgroup
}

\begin{document}

\title{\modelname{}: Tabular-Image Pre-training for Multimodal Classification with Incomplete Data} 

\titlerunning{\modelname{}: Tabular-Image Pre-training}

\author{Siyi Du$^{\dagger}$ \and
Shaoming Zheng \and
Yinsong Wang \and
Wenjia Bai \and 
Declan P. O'Regan \and 
Chen Qin$^{\dagger}$
}

\authorrunning{S. Du et al.}


\institute{Imperial College London, London, UK   \\
\email{\{s.du23,s.zheng22,y.wang23,w.bai,declan.oregan,c.qin15\}@imperial.ac.uk}}



\maketitle

\customfootnote{$^{\dagger}$Corresponding authors.}

\begin{abstract}
    Images and structured tables are essential parts of real-world databases. Though tabular-image representation learning is promising for creating new insights, it remains a challenging task, as tabular data is typically heterogeneous and incomplete, presenting significant modality disparities with images. Earlier works have mainly focused on simple modality fusion strategies in complete data scenarios, without considering the missing data issue, and thus are limited in practice. In this paper, we propose \modelname{}, a novel tabular-image pre-training framework for learning multimodal representations robust to incomplete tabular data. Specifically, \modelname{} investigates a novel self-supervised learning (SSL) strategy, including a masked tabular reconstruction task to tackle data missingness, and image-tabular matching and contrastive learning objectives to capture multimodal information. Moreover, \modelname{} proposes a versatile tabular encoder tailored for incomplete, heterogeneous tabular data and a multimodal interaction module for inter-modality representation learning. Experiments are performed on downstream multimodal classification tasks using both natural and medical image datasets. The results show that \modelname{} outperforms state-of-the-art supervised/SSL image/multimodal methods in both complete and incomplete data scenarios. Our code is available at \url{https://github.com/siyi-wind/TIP}.
    \keywords{Multimodal \and Image-tabular Representation Learning \and Missing Data \and Self-supervised Learning}
\end{abstract}

\section{Introduction}\label{sec:intro}
\begin{figure}[tb]
  \centering
  \includegraphics[width=1\linewidth]{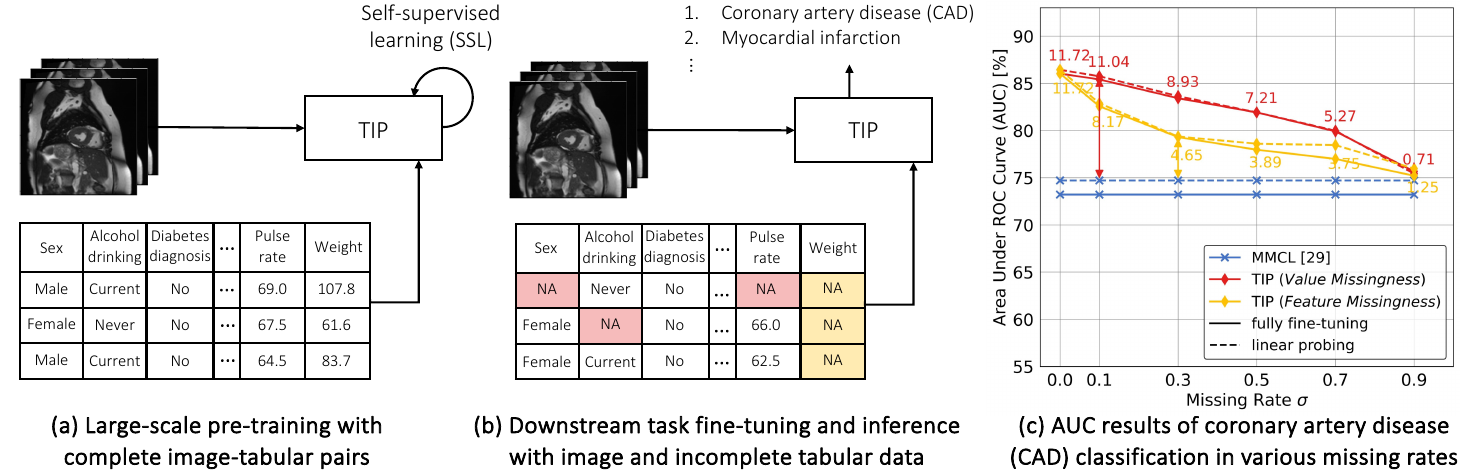}
  \caption{The pipeline of \modelname{}, which is pre-trained on large multimodal datasets (a) and can be deployed to downstream tasks with data missingness (b), \eg, \emph{Value Missingness} (red) and \emph{Feature Missingness} (yellow). Results for coronary artery disease classification (c) show \modelname{}'s superior performance over the SOTA multimodal pre-training method (numbers denote performance increase). Complete results in~\cref{fig:missingness}.
  }
  \label{fig:pipeline}
\end{figure}

While combining various modalities such as images and text to build a multimodal artificial intelligence (AI) system has achieved significant progress, integrating tabular data has been less explored~\cite{baltruvsaitis2018multimodal,bayoudh2022survey}. Tabular data, however, is increasingly accessible in multimodal datasets, and its integration is crucial in various applications~\cite{he2016ups,huang2022dvm,acosta2022multimodal,huang2020fusion,cai2019survey}. For instance, 
in healthcare, rich tabular information, \eg, demographics, lifestyle and laboratory tests (\cref{fig:pipeline}(a)), is commonly collected together with imaging data in hospitals, which are then used in a joint way to inform clinical decision-making~\cite{buntin2011benefits,chaudhry2006systematic,bai2020population}. Large population studies~\cite{bycroft2018uk,johnson2016mimic} such as the UK Biobank, have further enabled the wide availability of such multimodal resources to both machine learning and medical researchers. Despite these, current techniques for image-tabular data analysis are relatively limited. There is an increasingly strong interest in developing effective multimodal representation learning methods that can make the most of both image and tabular information to improve our understanding about human health.

Compared to vision-language modeling, incorporating image and tabular data in practical applications is a more difficult task with two main challenges. (1) Low-quality data: Despite the availability of large multimodal databases that allow pre-training, datasets for specific downstream tasks, \eg, classification of rare diseases, can be limited and often suffer from data sparsity~\cite{borisov2022deep}. In~\cref{fig:pipeline}(b), these datasets may inadvertently miss tabular values for some subjects, \ie, \emph{Value Missingness} (red table cells), or simply miss the entire features (columns), \ie, \emph{Feature Missingness} (yellow cells), due to diverse data collecting criteria across centers~\cite{barnard1999applications,jarrett2022hyperimpute,mackinnon2010use}. (2) Modality disparities: Unlike the homogeneous property of images and text, tabular data is heterogeneous with both dense numerical and sparse categorical features. The data columns exhibit varied value ranges, meanings and without clearly-defined inter-relationships~\cite{borisov2022deep}. Therefore, how to design a model that can effectively learn tabular and image representations to bridge the modality gap and address missing data is non-trivial.


Though there are some solutions for handling noisy and missing data~\cite{huang2020tabtransformer,somepalli2021saint,ghorbani2018embedding}, they mainly focus on unimodal tabular data analysis instead of multimodal tasks. Previous image-tabular models~\cite{wolf2022daft,zheng2022multi,huang2020fusion} are mostly trained and tested on relatively small labeled datasets (\eg, 653 samples~\cite{wolf2022daft}) with a limited amount of tabular information (\eg, 12 features~\cite{zheng2022multi}). These methods typically adopt shallow multi-layer perceptrons (MLPs) with simple modality fusion strategies and do not consider the challenges of incomplete data and modality disparities~\cite{huang2020tabtransformer}. Recently, Hager \etal~\cite{hager2023best}
proposed MMCL, the first SSL method that jointly trains image and tabular encoders through multimodal contrastive learning. However, it only utilizes the image encoder for downstream tasks, neglecting the wealth of information in the tabular data for decision-making.

In this work, to address the above-identified two challenges, we propose \modelname{}, a tabular-image pre-training framework based on a new multimodal representation learning network and a novel SSL pre-training strategy for managing small, incomplete downstream data (\cref{fig:pipeline}(a,b)). Specifically, we introduce a transformer-based tabular encoder with a versatile tabular embedding module, which serves two purposes: (1) it supports heterogeneous tabular inputs and diverse data missingness; (2) it captures inter-dependencies of tabular features and enhances representation learning. We further design a multimodal interaction module based on cross-modal attention to extract inter-modality information. To learn representations robust to missing data, we introduce three pre-training tasks: (1) masked tabular reconstruction to extract intra- and inter-modality relations from randomly masked data; (2) image-tabular contrastive learning to improve unimodal and multimodal representation learning; (3) image-tabular matching to obtain joint image-tabular representations for downstream tasks. Experiments on two representative datasets, \ie, cardiac data from the UK Biobank~\cite{bycroft2018uk} and natural image data from the DVM car advertisement dataset~\cite{huang2022dvm}, demonstrate \modelname{}'s superior performance. In particular, as illustrated in~\cref{fig:pipeline}(c), even with 50\% tabular values missing, \modelname{} has 7.21\% higher AUC than MMCL, the SOTA multimodal pre-training method.

Our contributions can be summarized as follows. (1) To the best of our knowledge, we are the first to propose SSL image-tabular pre-training to tackle the challenges of low-quality data and modality disparities and investigate various data missingness in multimodal tasks. (2) We introduce \modelname{}, featuring a transformer-based multimodal architecture for enhanced representation learning and a novel SSL pre-training strategy for tackling tabular missingness. (3) Experiments on both natural and medical datasets demonstrate that \modelname{} substantially surpasses SOTA supervised/SSL image/multimodal methods in both complete and incomplete data scenarios with various missing rates.

\section{Related Work}
\noindent\textbf{Self-supervised Learning (SSL)} approaches aim to acquire useful intermediate representations by pre-training models on unlabeled datasets with various intra- or inter-modal pretext tasks~\cite{jing2020self,min2023recent}. Two groups of intra-modal tasks are currently popular: (1) contrastive learning that models similarity (and dissimilarities) between multiple input views~\cite{chen2020simple,chen2021exploring,bahri2022scarf,wang2022transtab,somepalli2021saint}; and (2) generation-based learning that predicts the values of missing/corrupted input~\cite{he2022masked,assran2023self,devlin2019bert,brown2020language,ouyang2022training,yoon2020vime,ye2023ct,ucar2021subtab}. With the increasing availability of multimodal datasets, inter-modal pretext tasks are gaining more attention and have demonstrated remarkable performance~\cite{chen2023vlp,han2023survey}. Radford \etal~\cite{radford2021learning} introduced CLIP, which performs image-text contrastive pre-training on massive web data and exhibits notable zero-shot performance. Follow-up research~\cite{li2021align,li2022blip,yu2022coca} added tasks such as masked language modeling for more intricate cross-modal interactions. Nevertheless, few works have explored image-tabular pre-training~\cite{antelmi2021combining,hager2023best,ko2022deep}. Some generative image-tabular models~\cite{ko2022deep,antelmi2021combining} were proposed, but were limited to using two or four tabular features. Though MMCL~\cite{hager2023best} used 117 tabular features, it only supported unimodal downstream tasks, ignoring the usefulness of multimodal information in fine-tuning and inference time. We are the first to handle multimodal downstream tasks with incomplete data using tabular-image SSL pre-training.


\noindent\textbf{Deep Learning (DL) with Tabular Data}
has gained much interest~\cite{borisov2022deep}, due to their ability to achieve an end-to-end multimodal data learning~\cite{hager2023best,borisov2022deep}. Most existing works rely on MLPs and perform SSL tasks to learn representations~\cite{yoon2020vime,bahri2022scarf,ucar2021subtab}. Recently, transformers have been introduced to handle more challenging cases~\cite{majmundar2022met}, \eg, missing and noisy data~\cite{huang2020tabtransformer}, column permutation bias~\cite{yang2022tableformer}, and cross-table learning~\cite{wang2022transtab,ye2023ct}. These recent developments have inspired us to adapt the powerful transformer architecture for image-tabular learning.

\noindent\textbf{Multimodal Image-Tabular Learning} exploits tabular data as a complement to facilitate visual task learning, which is especially popular in the medical field~\cite{huang2020fusion,heiliger2023beyond,bayasi2024continual} and has achieved improved results compared to pure image models~\cite{polsterl2021combining,vale2021long}. Previous works typically extracted image and tabular features through two separate encoders that are fused through various methods~\cite{vale2021long,wolf2022daft,duanmu2020prediction,zheng2022multi,spasov2019parameter,duanmu2020prediction,borsos2024predicting}. However, these methods mostly tested on small datasets with limited tabular features and did not consider missing data. Some works~\cite{hager2023best,jiang2023transferring} transferred tabular knowledge into image models, but used image features alone for downstream tasks, ignoring potentially helpful information in tabular data.

\noindent\textbf{Missing Tabular Data} is common problem in scientific data analysis, for which many solutions have been proposed~\cite{miao2022experimental,schafer2002missing}. Some statistical approaches fill in missing values using column-wise mean or median~\cite{miao2022experimental}. An alternative popular method is iterative imputation, where each column with missing values is modeled as a function of other columns, employing a round-robin imputation process until convergence~\cite{jarrett2022hyperimpute,stekhoven2012missforest,raghunathan2001multivariate,royston2011multiple}. With the emergence of DL, imputation algorithms based on deep generative models were introduced~\cite{yoon2018gain,mattei2019miwae}, although they are limited to pre-processing steps and only support \emph{Value Missingness}~\cite{ghorbani2018embedding}. Some algorithms utilized SSL pre-training to make the model more robust to noisy or incomplete tabular data through reconstruction~\cite{yoon2020vime,huang2020tabtransformer}, contrastive learning~\cite{bahri2022scarf}, or denoising~\cite{somepalli2021saint}. In contrast, our study is the first to investigate various tabular missingness in a multimodal setting.

\section{Methodology}
\begin{figure}[tb]
  \centering
  \includegraphics[width=1\linewidth]{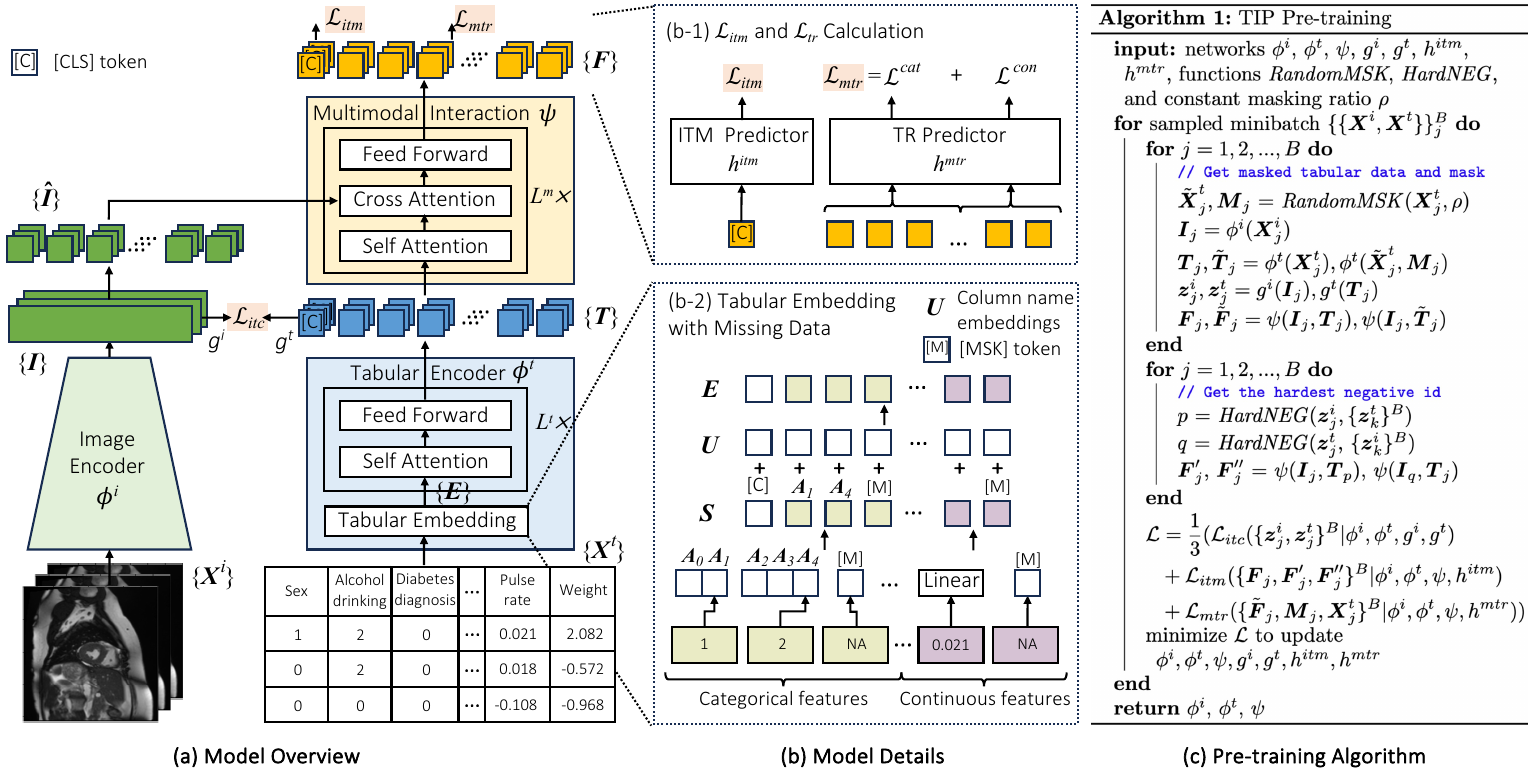}
  \caption{Model architecture and algorithm of \modelname{}: (a) Model overview with its image encoder, tabular encoder, and multimodal interaction module, which are pre-trained using 3 SSL losses: $\mathcal{L}_{itc}$, $\mathcal{L}_{itm}$, and $\mathcal{L}_{mtr}$. (b) Model details for (b-1) $\mathcal{L}_{itm}$ and $\mathcal{L}_{mtr}$ calculation and (b-2) tabular embedding with missing data. (c) Pre-training algorithm.
  }
  \label{fig:model}
\end{figure}

In this section, we introduce our \modelname{}, a tabular-image pre-training framework that is pre-trained on large multimodal datasets and then fine-tuned on downstream tasks, \eg, classification with complete/incomplete data. To encode incomplete, heterogeneous tabular data and enhance representation learning, we propose a tailored tabular encoder with versatile tabular embedding and transformer layers and a multimodal interaction module based on cross-modal attention. Moreover, we design a novel SSL pre-training strategy for multimodal information extraction and for addressing potential data missingness. The overall framework of \modelname{} is shown in~\cref{fig:model}. We describe \modelname{}'s model architecture in \cref{sec:model} and then discuss its SSL pre-training strategy in \cref{sec:pretraining}.

\subsection{\modelname{} Model Architecture}\label{sec:model}
Let $(\boldsymbol{X}^i \in \mathbb{R}^{H \times W \times 3}, \boldsymbol{X}^t=[x^t_1,...,x^t_N] \in \mathbb{R}^N)$ be an image-tabular pair, where $N$ is the number of tabular features. Assume that each tabular input contains $N_a$ categorical features, $x^t_1, ..., x^t_{N_a}$, and $N-N_a$ continuous features, $x^t_{N_a+1}, ..., x^t_{N}$. We convert categorical data into ordinal numbers and standardize continuous data as~\cite{hager2023best}. As shown in~\cref{fig:model}(a), \modelname{} involves a convolutional neural network (CNN) based image encoder $\phi^i$, a tabular encoder $\phi^t$, and a multimodal interaction module $\psi$. The image representation $\boldsymbol{I} \in \mathbb{R}^{H' \times W' \times C}$ and the tabular representation $\boldsymbol{T} \in \mathbb{R}^{N \times D}$ are extracted by the image and tabular encoders, respectively, where $C$ and $D$ are their corresponding channel dimensions. We transform and project $\boldsymbol{I}$ into a sequence of embeddings $\boldsymbol{\hat{I}} \in \mathbb{R}^{(H'W') \times D}$. Based on that, the multimodal interaction module then receives input of $\boldsymbol{\hat{I}}$ and $\boldsymbol{T}$ to perform inter-modality learning, yielding a multimodal representation $\boldsymbol{F} \in \mathbb{R}^{N \times D}$.

\noindent\textbf{Tabular Encoder:} To tackle heterogeneous data with potential missingness and to extract rich contextual information from tabular features, we propose to treat each tabular feature as a basic element and convert it to a token embedding through a versatile tabular embedding module (\cref{fig:model}(b-2)). We then enable the embedded tokens to attend to related tokens through transformer layers. Our tabular embedding module contains 3 parts: \emph{heterogeneous data processing}, \emph{missing data processing}, and \emph{column diversity integration}.

(1) \emph{Heterogeneous data processing:} Tabular data usually contains categorical and continuous variables, which are very different attribute types and cannot be embedded using a single function~\cite{borisov2022deep,gorishniy2021revisiting}. Therefore, our \emph{tabular embedding} module processes these two types of features independently. In particular, we convert each categorical feature into a token embedding using a learnable embedding matrix $\boldsymbol{A} \in \mathbb{R}^{\bar{N}_{a} \times D}$, where $\bar{N}_{a}$ is a summation of the number of unique values in $N_a$ categorical feature. Meanwhile, We employ a shared linear layer to project each continuous feature into the D-dimensional space. 
 
(2) \emph{Missing data processing:} To handle incomplete data and inform our model which data is missing during training, we propose to embed each missing value using a special trainable $D$-dimentional [MSK] token embedding. This method does not require data imputation before inputting and can handle various types of missing data scenarios, which is more flexible than previous techniques~\cite{jarrett2022hyperimpute,yoon2020vime,bahri2022scarf} that only support random \emph{Value Missingness}~\cite{ghorbani2018embedding}. The embedded features are concatenated with a special tunable [CLS] token to generate a sequence of embeddings $\boldsymbol{S}$. The [CLS] token's state at the last transformer layer serves as a learnt representation for downstream classification tasks as in~\cite{li2021align,li2022blip}.

(3) \emph{Column diversity integration:} Each column in tabular data generally has a different meaning, thus it is not suitable to treat them uniformly as pixels in images~\cite{ko2022deep,yang2022tableformer}. To distinguish different columns and capture inter-column relationship, we propose to integrate column diversity through a sequence of learnable column name embeddings $\boldsymbol{U} \in \mathbb{R}^{(N+1) \times D}$. Instead of using pre-trained language models to tokenize column names into fixed textual embeddings~\cite{ye2023ct,wang2022transtab,yang2022tableformer}, our data-driven strategy can dynamically capture hidden column dependencies existing in the training data. For example, the `weight' and `alcohol drinking' columns may have little semantic similarity but can present strong clinical association. The final embeddings are formulated as: $\boldsymbol{E} = \boldsymbol{S}+\boldsymbol{U}$.

Driven by transformers' ability to capture long-range dependencies through self-attention~\cite{vaswani2017attention} and to embed knowledge from large databases~\cite{han2023survey,kalyan2022ammu}, we utilize $L_t$ transformer layers~\cite{vaswani2017attention} to encode tabular information and extract a high-level tabular representation $\boldsymbol{T}$. To avoid the potential negative effect of missing data on model learning~\cite{he2022masked}, we employ a self-attention mask, which restricts each token to only attend to itself and non-missing tokens, thus ensuring the tabular representation learning to be more robust and stable.

\noindent\textbf{Multimodal Interaction Module:} To enhance multimodal representation learning and capture cross-modal relationships, we propose to leverage the cross-attention mechanism in a transformer decoding module~\cite{vaswani2017attention} to enable the [CLS] token and each tabular token to cross-attend to relevant image information. The interaction module consists of $L^m$ layers, each including self-attention, cross-modal attention, an MLP feed-forward module, and layer normalization. The cross-modal attention in the $l$th layer can be written as:
\begin{equation}\label{equation:crossatt}
CrossAttention(\boldsymbol{Q},\boldsymbol{K},\boldsymbol{V}) = softmax(\boldsymbol{Q}\boldsymbol{K}^T/\sqrt{d_k})\boldsymbol{V},
\end{equation}
\noindent where $\boldsymbol{Q}=\boldsymbol{F}_{l-1} \boldsymbol{W}^Q_l$, $\boldsymbol{K}=\hat{\boldsymbol{I}}\boldsymbol{W}^K_l$, $\boldsymbol{V}=\hat{\boldsymbol{I}}\boldsymbol{W}^V_l$, and $\boldsymbol{F}_0=\boldsymbol{T}$.

\subsection{SSL Pre-training Strategy}\label{sec:pretraining}
To enable the model to be robust to incomplete downstream data while improving representation learning, we propose to pre-train our model with three objectives, including masked tabular reconstruction, image-tabular contrastive learning, and image-tabular matching, as shown in~\cref{fig:model}(a,c).

\noindent\textbf{Image-Tabular Contrastive Learning (ITC):} We design the ITC task to capture better unimodal representations and align their feature spaces before modality fusion. This shares a similar motivation with image-text contrastive learning, which has demonstrated the ability to extract transferable representations for downstream tasks and to facilitate cross-modal learning~\cite{li2022blip,li2021align,radford2021learning}. ITC encourages image and tabular representations from a matched sample to be close compared to those from unmatched ones. We utilize two projection heads $g^i$ and $g^t$ to bring $\boldsymbol{I}$ and $\boldsymbol{T}$ to a shared low-dimensional hidden space and calculate the image-to-tabular and tabular-to-image similarities, \ie, $s_{j}^{i2t}$ and $s_j^{t2i}$, as~\cite{li2021align}. The image-tabular contrastive loss can then be computed as $\mathcal{L}_{itc}=-(s_{j}^{i2t}+s_j^{t2i})/2$.

\noindent\textbf{Image-Tabular Matching (ITM):} We propose the ITM task to enable \modelname{}'s multimodal interaction module to capture inter-modality relations and generate a joint multimodal representation, inspired by the success of image-text matching~\cite{li2022blip,li2021align}. Our ITM aims to predict whether a pair of imaging and tabular input is positive (matched) or negative (unmatched). As displayed in~\cref{fig:model}(b-1), we feed the [CLS] embedding of $\boldsymbol{F}$, which captures a joint representation of an image-tabular pair, into the ITM predictor $h^{itm}$ (a linear layer) for matching prediction based on a binary cross-entropy loss $\mathcal{L}_{itm}$. To enhance representation learning and capture discriminative features, we expose the model to more informative negative pairs using the hard negative mining strategy (\emph{HardNEG}) proposed in~\cite{li2021align}. Specifically, for each image/tabular representation, we select one unmatched tabular/image representation from the mini-batch using the similarity calculated in ITC as the sampling weight.

\noindent\textbf{Masked Tabular Reconstruction (MTR):} This task aims to learn multimodal representations that can be robust to missing tabular data in downstream tasks. Previous studies have found that reconstructing masked data can help mitigate noisy or missing data issues and produce promising performance in SSL representation learning~\cite{zhang2023survey,dong2022table,huang2020tabtransformer,sun2022survey,pathak2016context,majmundar2022met}. We therefore propose MTR for multimodal representation learning and require the model to predict the masked parts in tabular data based on both image and unmasked tabular information. Specifically, we apply random masking (\emph{RandomMSK}) on a tabular input based on a masking ratio $\rho$, to generate a masked version $\tilde{\boldsymbol{X}}^t$, \ie, treating masked values as missing data, and a mask matrix $\boldsymbol{M}$ used to record masking positions. The resulting $\tilde{\boldsymbol{X}}^t$ is then fed into the model to produce the masked multimodal representation $\tilde{\boldsymbol{F}}=[\tilde{\boldsymbol{f}}_1,...,\tilde{\boldsymbol{f}}_N$]. The generated $\tilde{\boldsymbol{F}}$ serves as the input to a MTR predictor $h^{mtr}$ for reconstructing the missing values. Since reconstructing categorical data is a classification task, whereas reconstructing continuous data is a regression task, $h^{mtr}$ has two distinct linear layers, processing $N_a$ categorical and $N-N_a$ continuous features correspondingly:
\begin{equation}\label{equation:tr}
  h^{mtr}(\tilde{\boldsymbol{f}}_n) = \begin{cases}
      \boldsymbol{W}_1 \tilde{\boldsymbol{f}}_n + \boldsymbol{b}_1, \boldsymbol{W}_1 \in \mathbb{R}^{\bar{N}_a \times D}  & 0<n\leq N_a, \\
      \boldsymbol{W}_2\tilde{\boldsymbol{f}}_n+b_2, 
      \boldsymbol{W}_2 \in \mathbb{R}^{1 \times D} &  N_a<n\leq N.
  \end{cases}
\end{equation}

We formulate a reconstruction loss based on masked features only, $\mathcal{L}_{mtr}=\mathcal{L}^{cat}+\mathcal{L}^{con}$, where $\mathcal{L}^{cat}$ represents the cross-entropy loss for categorical features and $\mathcal{L}^{con}$ is a mean squared error loss for continuous features. Compared to previous tabular techniques that fill the masked cell with a randomly selected value from the same column~\cite{yoon2020vime,huang2020tabtransformer}, our MTR task with the random masking strategy enables the model to learn a mask token for missing data in downstream tasks, so that the model can fully understand which values are missing and handle diverse data missingness, even if a whole column is missing. Ultimately, the overall pre-training loss function is formulated as:
\begin{equation}\label{equation:loss}
   \mathcal{L} = (\mathcal{L}_{itc}+\mathcal{L}_{itm}+\mathcal{L}_{mtr})/3.
\end{equation}

\noindent\textbf{Ensemble Learning during Fine-tuning:} After pre-training, we can add linear classifiers after the feature extractor for downstream classification tasks. Given that our pre-training strategy enables the image encoder, tabular encoder, as well as multimodal interaction module to learn rich representations beneficial to downstream tasks,  as well as motivated by ensemble learning's ability to boost models' generalizability and results~\cite{dong2020survey,ganaie2022ensemble}, we propose to build an ensemble model to further enhance the model's performance. Specifically, we incorporate a linear classifier after each of the three modules and average the predictions from all three classifiers to generate the final output.

\section{Experiment}
\noindent\textbf{Datasets and Metrics:} Similar to~\cite{hager2023best}, we experiment on two large datasets: a medical dataset -- UK Biobank (UKBB)~\cite{bycroft2018uk} and a natural image dataset -- Data Visual Marketing (DVM)~\cite{huang2022dvm}. UKBB contains rich cardiac imaging and clinical tabular data gathered from individuals in the United Kingdom~\cite{littlejohns2020uk}. We carry out two cardiac disease classification tasks: coronary artery disease (CAD) and myocardial infarction (Infarction), using 2D short-axis cardiac magnetic resonance (MR) images and 75 disease-related tabular features (details in Sec. A of the supplementary material (supp.)). The dataset contains 36,167 image-tabular pairs, split into training (26,040),  validation (6,510), and test (3,617) sets. Due to the low disease prevalence (3\% for Infarction and 6\% for CAD), we use balanced training datasets for fine-tuning, comprising 3,482 for CAD and 1,552 for Infarction, and evaluate all models with area under the curve (AUC). DVM~\cite{huang2022dvm} is a publicly available dataset for automotive applications, including 2D car images and car-related tabular data. We obtain 176,414 image-tabular pairs (17 tabular features, details in Sec. A of supp.) and implement a car model classification task with 283 classes. We split this dataset into training (70,565), validation (17,642), and test (88,207) sets and use accuracy for evaluation.

\noindent\textbf{Implementation Details:} We utilized a ResNet-50~\cite{he2016deep} as the image encoder. Our tabular encoder and multimodal interaction module both have 4 transformer layers, with 8 attention heads and a hidden dimension of 512. We used an MLP with a hidden size of 2048 for $g^i$, and a hidden size of 512 for $g^t$ in ITC. Both MLPs have an output size of 128. The temperature parameter $\tau$ for ITC is 0.1, and the masking ratio $\rho$ for MTR is 0.5. The images are resized to $128 \times 128$. During pre-training, we conducted tabular augmentation for ITC and ITM, and image augmentation for 3 pre-training tasks. Note that the CAD and Infarction tasks utilize the same pre-trained model during fine-tuning. Additional implementation details for \modelname{} and other comparing models are in Sec. B of supp..

\begin{table}[tb]
  \caption{Results of DVM, CAD, and Infarction classification tasks comparing \modelname{} with supervised/SSL image/multimodal techniques on complete data. \faSnowflake[regular] denotes linear probing, \ie, the feature extractors are frozen, and only the linear classifiers of the pre-trained models are tuned. \faFire* means fully fine-tuning, \ie, all parameters are trainable. For supervised methods, all parameters are trainable in both \faSnowflake[regular] and \faFire* columns.
  }
  \label{tab:SOTA}
  \centering
  \resizebox{\textwidth}{!}{\begin{tabular}{p{35mm}|P{17mm}P{17mm}|P{17mm}P{17mm}|P{17mm}P{17mm}}
    \hline
    Model & \multicolumn{2}{c|}{DVM Accuracy (\%) $\uparrow$} & \multicolumn{2}{c|}{CAD AUC (\%) $\uparrow$} & \multicolumn{2}{c}{Infarction AUC (\%) $\uparrow$} \\
    \cline{2-7}
    ~ & \small\faSnowflake[regular] & \faFire* &  \faSnowflake[regular] & \faFire* &  \faSnowflake[regular] & \faFire* \\
    \hline
    \multicolumn{7}{c}{(a) Supervised Image and Multimodal Methods} \\ 
    \hline
    ResNet-50~\cite{he2016deep}  & 87.68  & 87.68 & 63.11 & 63.11  & 59.48 & 59.48 \\
    \hline
    Concat Fuse 
    (CF)~\cite{spasov2019parameter} & 94.60 & 94.60 & 85.76 & 85.76 & \textbf{85.05} & 85.04 \\
    Max Fuse (MF)~\cite{vale2021long}  & 94.39 & 94.39 & 85.31 & 85.31 & 84.75 & 84.75 \\
    Interact Fuse (IF)~\cite{duanmu2020prediction}  & 96.24 & 96.24 & 84.89 & 84.89 & 81.91 & 81.91 \\
    DAFT~\cite{wolf2022daft} & 96.60 & 96.60 & 86.21 & \textbf{86.21} & 56.27 & 56.27 \\
    \hline
    \multicolumn{7}{c}{(b) SSL Image Pre-training Methods} \\ 
    \hline
    SimCLR~\cite{chen2020simple} & 61.06 & 87.65 & 68.42 & 72.58 & 68.86 & 75.07 \\
    BYOL~\cite{grill2020bootstrap} & 56.26 & 88.64 & 65.67 & 69.18 & 66.63 & 70.12\\
    SimSiam~\cite{chen2021exploring} & 23.14 & 78.62 & 57.77 & 67.71 & 53.83 & 64.79 \\
    BarlowTwins~\cite{zbontar2021barlow} & 53.60 & 88.36 & 55.64 & 61.68 & 50.01 & 60.14 \\
    \hline 
    \multicolumn{7}{c}{(c) SSL Multimodal Pre-training Methods} \\ 
    \hline
    MMCL~\cite{hager2023best} & 91.66 & 93.27 & 74.71 & 73.21 & 76.79 & 76.46 \\
    \rowcolor{gray!25}
    TIP & \textbf{99.72} & \textbf{99.56} & \textbf{86.43} & 86.03 & 84.46 & \textbf{85.58} \\
  \hline
  \end{tabular}}
\end{table}

\subsection{Comparison with SOTAs on Complete Downstream Data}\label{sec:complete data}
We first investigate the performance of \modelname{} in a complete downstream data regime by comparing it with other supervised and SSL pre-training algorithms. For supervised learning, we trained a fully supervised image model, ResNet-50, and reproduced 4 image-tabular learning strategies: concatenation fusion (CF)~\cite{spasov2019parameter}, maximum fusion (MF)~\cite{vale2021long}, interactive fusion through channel-wise multiplication (IF)~\cite{duanmu2020prediction}, and dynamic affine feature map transform (DAFT)~\cite{wolf2022daft}. For fair comparison, the image encoder used in all these approaches is ResNet-50. For SSL image pre-training, we tested 4 popular contrastive learning solutions: SimCLR~\cite{chen2020simple}, BYOL~\cite{grill2020bootstrap}, SimSiam~\cite{chen2021exploring}, and BarlowTwins~\cite{zbontar2021barlow}. We also compared our \modelname{} with MMCL~\cite{hager2023best}, a recent multimodal image-tabular pre-training method. We evaluated all pre-trained models using linear probing, which only tunes linear classifiers, and fully fine-tuning, which trains all parameters.

As shown in~\cref{tab:SOTA}(a,b), \modelname{} outperforms supervised/SSL image-only models in linear probing and fully fine-tuning by a large margin, which indicates that integrating multiple modalities in pre-training improves the representation learning and that tabular information facilitates our classification tasks. Moreover, \modelname{} significantly surpasses MMCL, \eg, in linear probing, boosting the accuracy by 8.06\% on DVM and AUC by 11.72\% on CAD. While MMCL transfers tabular information related to visual features into the image branch and discards the tabular branch during fine-tuning, tabular data often contains task-related complementary information that is not visible in images~\cite{acosta2022multimodal,bai2020population}. Our results showcase \modelname{} can exploit tabular information that is visible or non-visible in images to improve downstream tasks. Finally, compared to supervised multimodal methods (\cref{tab:SOTA}(a)), \modelname{} achieves higher performance on DVM, \eg, raising accuracy by 3.12\% in linear probing. On CAD and Infarction tasks, \modelname{} performs competitively against supervised multimodal methods, indicating the usefulness of features learnt via self-supervised pre-training and a requirement for a larger pre-training dataset (70,565 in DVM \vs 26,040 in UKBB).

\begin{figure}[tb]
  \centering
  \includegraphics[width=1\linewidth]{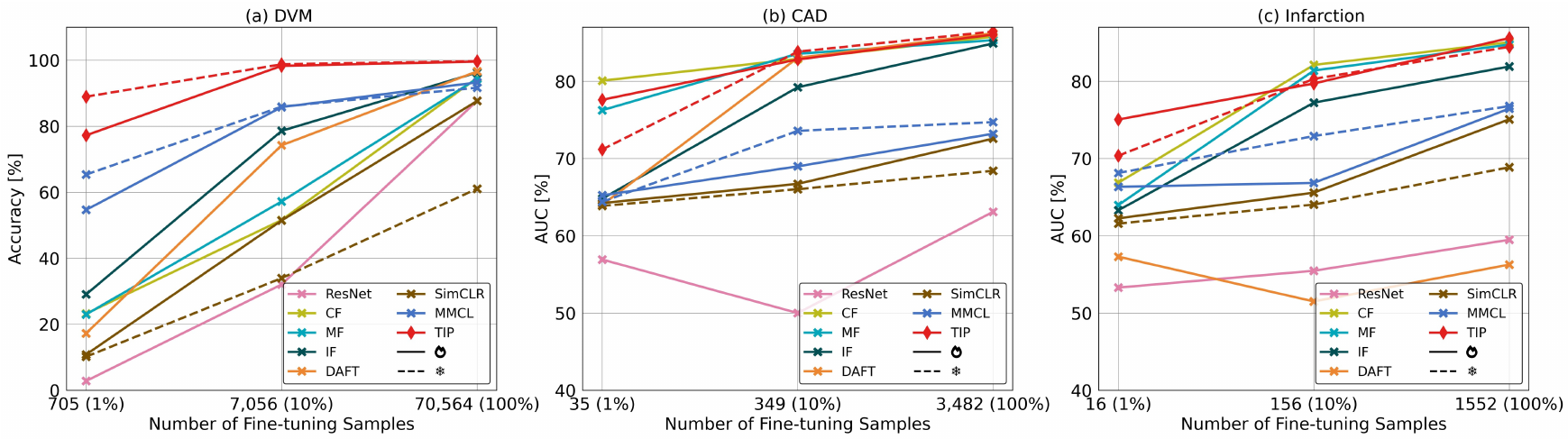}
  \caption{Result comparison with supervised/SSL image/multimodal methods on various number of fine-tuning samples. \faFire* denotes fully fine-tuning, and \faSnowflake[regular] means linear probing.
  }
  \label{fig:low}
\end{figure}

\noindent\textbf{Robustness to Low-data Regimes:}
As data annotation for downstream tasks is often costly, we propose to assess the performance of \modelname{} and other SOTA methods on low-data regimes (10\% and 1\% of the original training data size). We used 7,056 (10\%) and 705 (1\%) training samples for DVM, 349 and 35 for CAD, and 156 and 16 for Infarction. For SSL image approaches, only SimCLR's results are presented since it showed the best performance among them (complete results in Sec. C of supp.). \cref{fig:low} shows that \modelname{} is more robust at low-data regimes and outperforms other SOTAs on DVM. For \modelname{}, 10\% of the training data can already achieve a performance close to that of 100\%, indicating the potential to use fewer data for fine-tuning. Only for two cases, 1\% CAD and 10\% Infarction, \modelname{} slightly underperforms CF, a supervised multimodal method, possibly due to the relatively small pre-training datasets of CAD and Infarction used by \modelname{}.

\begin{figure}[tb]
  \centering
  \includegraphics[width=1\linewidth]{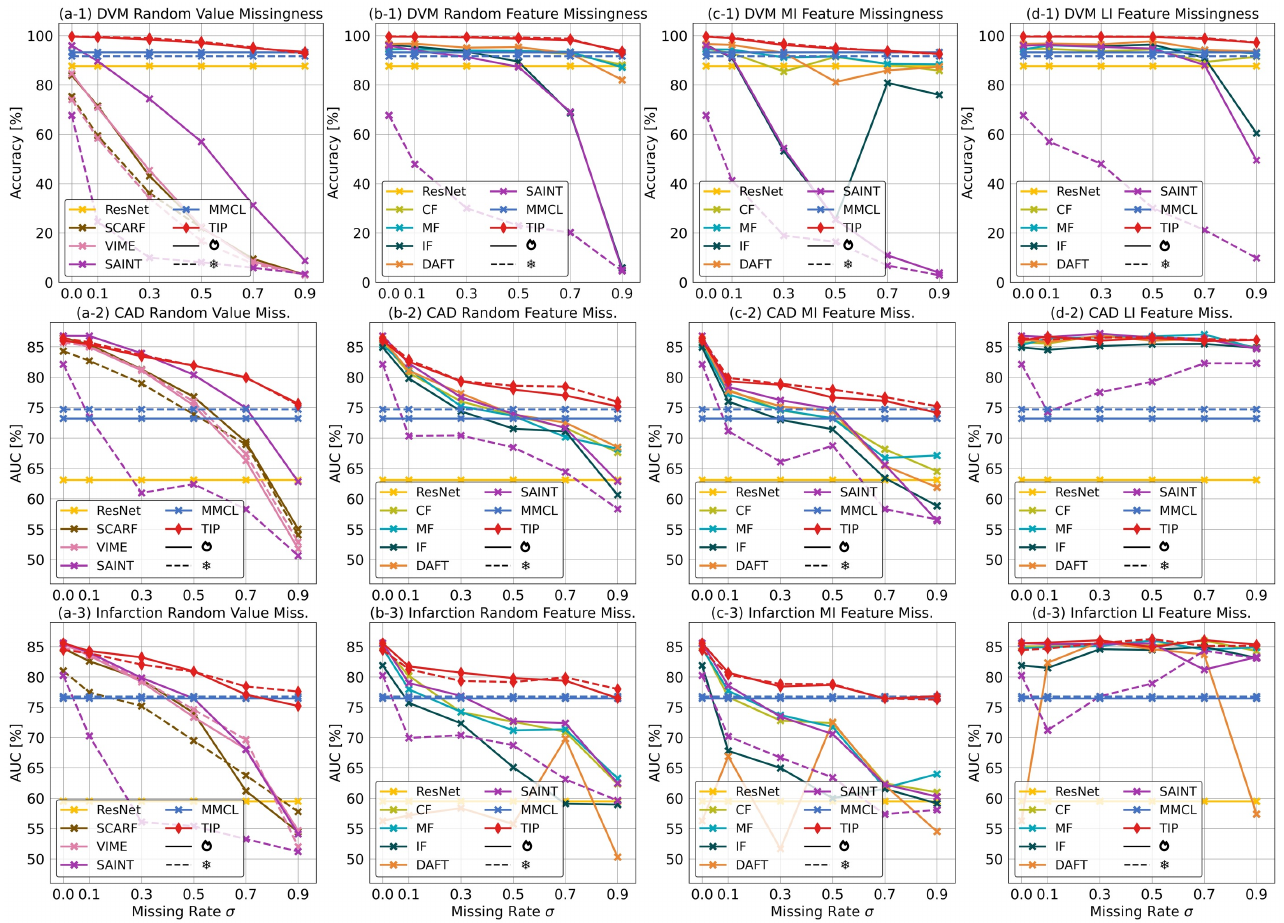}
  \caption{Results of 4 missing scenarios: (a) RVM, (b) RFM, (c) MIFM, and (d) LIFM, on DVM (1st row), CAD (2nd row), and Infarction (3rd row) tasks with different missing rates $\sigma$. \faFire* denotes fully fine-tuning, and \faSnowflake[regular] means linear probing.
  }
  \label{fig:missingness}
\end{figure}

\subsection{Comparison with SOTAs on Incomplete Downstream Data}
We conduct a study to assess the model performance on tackling tabular missingness. To achieve that, we introduce 4 types of missing scenarios: (a) random value missingness (RVM), where tabular values (cells) are randomly missing; (b) random feature missingness (RFM), where a random set of features (columns) is missing; (c) most important feature missingness (MIFM), where the most important features for the prediction task are eliminated in descending order; (d) least important feature missingness (LIFM), where the least important features are removed first. The importance of features is determined using a random forest algorithm~\cite{liaw2002classification} trained on downstream training datasets. To showcase \modelname{}'s capability in handling missing data using multimodal information, we compared it with 3 SSL tabular pre-training methods: VIME~\cite{yoon2020vime}, SCARF~\cite{bahri2022scarf}, and SAINT~\cite{somepalli2021saint}. MLP-based VIME and SCARF do not support feature missingness, while transformer-based SAINT can handle all missing scenarios. Hence, we only evaluated VIME and SCARF in RVM by filling missing positions with randomly chosen values from the same column, as in~\cite{yoon2020vime,bahri2022scarf}. Supervised multimodal approaches cannot address random value missingness. Thus, they are compared in RFM, MIFM, and LIFM. SSL image techniques and MMCL are not affected by missing tabular data. For complete comparison, we include the highest MMCL's result among them. We executed each scenario with 6 missing rates.



\begin{table}[tb]
  \caption{Result comparison of \modelname{} and data imputation methods for reconstructing missing continuous features across various missing rates on DVM and UKBB test sets.
  }
  \label{tab:imputation}
  \centering
  \resizebox{\textwidth}{!}{\begin{tabular}{p{27mm}|P{17mm}P{17mm}P{17mm}|P{17mm}P{17mm}P{17mm}}
  \hline
  Model & \multicolumn{3}{c|}{DVM RMSE $\downarrow$} & \multicolumn{3}{c}{UKBB RMSE $\downarrow$} \\
  \cline{1-7}
  Missing rate $\sigma$ & 0.3 & 0.5 & 0.7 & 0.3 & 0.5 & 0.7 \\
  \hline
  Mean~\cite{hawthorne2005imputing} & 0.9621 & 0.9783 & 0.9733 & 1.0162 & 1.0191 & 1.0070 \\
  MissForest~\cite{stekhoven2012missforest} & 0.6700 & 0.7653 & 0.8833 & 0.7516 & 0.7754 & 0.8177 \\
  GAIN~\cite{yoon2018gain} & 1.0447 & 0.9428 & 2.9705 & 0.7920 & 2.0039 & 2.8130 \\
  MIWAE~\cite{mattei2019miwae} & 1.0105 & 1.0265 & 1.0218 & 1.0644 & 1.0680 & 1.0557 \\
  Hyperimpute~\cite{jarrett2022hyperimpute} & 0.6329 & 0.9428 & 0.9793 & 0.6803  & 0.7242 & 0.8060  \\
  \rowcolor{gray!25}
  TIP & \textbf{0.3899} & \textbf{0.4651} & \textbf{0.5055} & \textbf{0.6039} & \textbf{0.6460} & \textbf{0.7106} \\
  \hline
  \end{tabular}}
\end{table}

As depicted in~\cref{fig:missingness}, the most challenging scenario is MIFM, where most models' performance drops significantly. Besides, supervised multimodal methods outperform ResNet and MMCL in MIFM and RFM when $\sigma=0.1$ and achieve improved results across various missing rates in LIFM. This indicates even if downstream data is incomplete, they can still provide useful information, especially when MI features are not missing. We notice that some models show an increase with higher missing rates in MIFM and attribute this to their feature importance not being exactly the same as identified by the random forest model.

In comparison to other approaches, \modelname{} can cope with all types of data missingness scenarios and significantly surpasses other methods. For supervised multimodal models (\cref{fig:missingness}(b,c,d)), missing data can significantly reduce their performance, especially when the missing rate is high. However, \modelname{} remains robust across different missing rates and exhibits improved performance. For instance, in RFM ($\sigma=0.5$), \modelname{} increases accuracy by 3.9\% on DVM and AUC by 4.7\% on CAD compared to DAFT. This implies that our pre-training strategy allows the model to capture valuable multimodal embeddings and their relationships from unlabeled image-tabular pairs. 

Moreover, \modelname{} performs better than tabular pre-training techniques (VIME and SCARF in~\cref{fig:missingness}(a)), especially with a high missing rate, \eg, it increases accuracy by 75.02\% on DVM when $\sigma=0.5$. When compared to SAINT, which also uses learnable mask tokens during fine-tuning, \modelname{}'s superior performance indicates that our SSL strategy is more effective for incomplete downstream data and allows the model to be able to integrate visual and tabular information for predicting missing components. Finally, compared to the multimodal pre-training model MMCL, \modelname{} outperforms it by a large margin, even with a high missing rate of 0.7, in all scenarios. This shows that \modelname{} can fully leverage tabular information in incomplete data. Even when certain tabular features in downstream tasks are inaccessible, the intra- and inter-modality relations learnt during pre-training enable \modelname{} to generate promising outcomes.

\noindent\textbf{Missing Value Reconstruction:} We further assess \modelname{}'s missing data reconstruction performance by comparing with 5 data imputation methods: column-wise mean substitution (Mean)~\cite{hawthorne2005imputing}, MissForest~\cite{stekhoven2012missforest}, GAIN~\cite{yoon2018gain}, MIWAE~\cite{mattei2019miwae}, and HyperImpute~\cite{jarrett2022hyperimpute}. Since GAIN and MIWAE are hard to apply to mixed-type data containing both continuous and categorical features, our experiments focus on the continuous data within DVM and UKBB test sets, using root mean squared error (RMSE) for evaluation. In this case, we masked the categorical data input for \modelname{}. As shown in~\cref{tab:imputation}, \modelname{} exceeds those imputation algorithms across varying missing rates, which showcases that our pre-training task and multimodal architecture have enabled the model to capture the relations with multimodal data and thus predict missing information more accurately.

\begin{table}[tb]
  \caption{Experiments using different image encoder backbones and ablation study of \modelname{}. \faSnowflake[regular] means linear probing, and \faFire* represents fully fine-tuning.
  }
  \label{tab:ablation}
  \centering
  \resizebox{\textwidth}{!}{\begin{tabular}{p{40mm}|P{16mm}P{16mm}|P{16mm}P{16mm}|P{16mm}P{16mm}}
    \hline
    Model & \multicolumn{2}{c|}{DVM Accuracy (\%) $\uparrow$} & \multicolumn{2}{c|}{CAD AUC (\%) $\uparrow$} & \multicolumn{2}{c}{Infarction AUC (\%) $\uparrow$} \\
    \cline{2-7}
    ~ & \small\faSnowflake[regular] & \faFire* &  \faSnowflake[regular] & \faFire* &  \faSnowflake[regular] & \faFire* \\
    \hline
    \multicolumn{7}{c}{(a) Applicability to Various Image Encoder Backbones} \\ 
    \hline
    TIP (ViT-S~\cite{dosovitskiy2020image}) & 99.67 & 99.40 & 85.85 & 86.94 & 83.83 & 86.16 \\
    TIP (ViT-B~\cite{dosovitskiy2020image}) & 99.40 & 99.28 & 84.90 & 86.93 & 83.15 & 85.76 \\
    \hline
    \multicolumn{7}{c}{(b) Ablation Study} \\ 
    \hline
    TIP w/o SSL pre-training & 98.57 & 98.57 & 86.04 & 86.04 & 84.19 & 84.19\\
    TIP w/o column name emb. & 97.38 & 97.21 & 79.40 & 81.12 & 82.00 & 75.15 \\
    TIP w/o ensemble & 99.63 & 99.35 & 86.00 & \textbf{86.97} & 84.43 & 84.00 \\
    \rowcolor{gray!25}
    TIP & \textbf{99.72} & \textbf{99.56} & \textbf{86.43} & 86.03 & \textbf{84.46} & \textbf{85.58} \\
  \hline
  \end{tabular}}
\end{table}


\begin{table}[!t]
\begin{minipage}[t]{0.53\textwidth}
	\centering
        \captionsetup{type=figure}
	\includegraphics[width=1.0\textwidth]{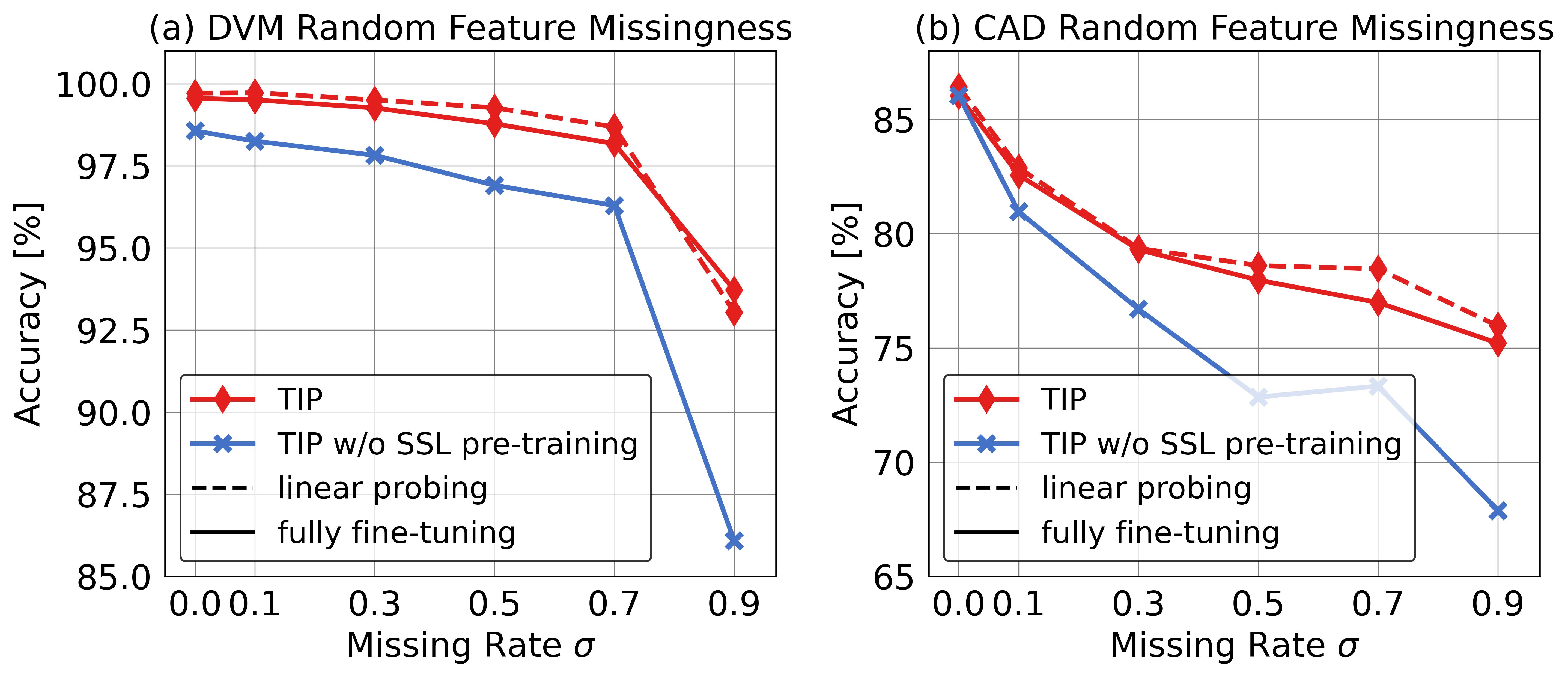}
	    \caption
	{
            \fontsize{9}{9}\selectfont
		Results comparing \modelname{} with or without the proposed SSL pre-training strategy on the DVM and the CAD RFM scenario.
	}
	\label{fig:SSL}
\end{minipage}
\hfill
\begin{minipage}[t]{0.45\textwidth}
     \caption
	{
            \fontsize{9}{9}\selectfont
		\modelname{}'s RMSE results on the DVM test set for missing continuous feature reconstruction. $\sigma$ denotes missing rate, and $\rho$ means masking ratio.
	}
        \fontsize{9}{9}\selectfont
	\centering	
     \setlength\tabcolsep{4pt}
		\begin{tabular}	{l  |  c c c }
			\hline
			$\sigma$ & 0.3 & 0.5 & 0.7 \\
			\hline
                $\rho=0.1$ & .5349 & .6752 & .7871 \\
			$\rho=0.3$ & .4110 & .5128 & .5924 \\
			$\rho=0.5$ & \textbf{.3899} & .4651 & .5055 \\
			$\rho=0.7$ & .3986 & \textbf{.4612} & \textbf{.4733} \\	
                $\rho=0.9$ & .4279 & .4800 & .4816 \\	
			\hline
		\end{tabular}
    \fontsize{9}{9}\selectfont
	\label{tab:masking rmse}	
\end{minipage}
\end{table}

\subsection{Ablation Study and Visualization}
\noindent\textbf{Applicability to Different Image Encoder Backbones:} We propose to vary the image encoder backbone to showcase the general applicability of the proposed method. Specifically, we utilized two vision transformers (ViTs): ViT-S/16 and ViT-B/16~\cite{dosovitskiy2020image}, as the variations for the image encoder. Since ViTs output a sequence representation, we directly project it into the same hidden dimension as the tabular representation. \cref{tab:ablation}(a) exemplifies that using ViTs results in similar outcomes to using ResNet-50, and large ViT-B does not perform better than small ViT-S. We suspect this is due to ViTs performing better than CNNs when pre-trained on much larger datasets, as found in~\cite{dosovitskiy2020image}.

\noindent\textbf{Ablation Study on Key Model Components:} We directly trained \modelname{}'s model architecture on downstream tasks in a supervised manner to evaluate the efficacy of our SSL pre-training strategy. \cref{tab:ablation}(b) and \cref{fig:SSL} demonstrate that our pre-traning strategy improves the performance of downstream tasks, especially on incomplete data with a high missing rate. In addition, we conducted experiments that removed the column name embeddings in our tabular encoder or the ensemble learning during fine-tuning. \cref{tab:ablation}(b) shows that subtracting any of those techniques leads to inferior performance. Additional ablation studies on each pre-training task and \modelname{}'s tabular encoder in Sec. C.2-3 of supp..

\noindent\textbf{Sensitivity Analysis of the Masking Ratio:} We study the impact of different masking ratios $\rho$ in the MTR task. \cref{tab:masking rmse} shows that moderate masking ratios $\rho \in {(0.5, 0.7)}$ achieve the best performance, whereas too high (0.9) or too low (0.1) ratios adversely affect model learning. More analysis in Sec C.4 of supp..

\begin{figure}[tb]
  \centering
  \includegraphics[width=1\linewidth]{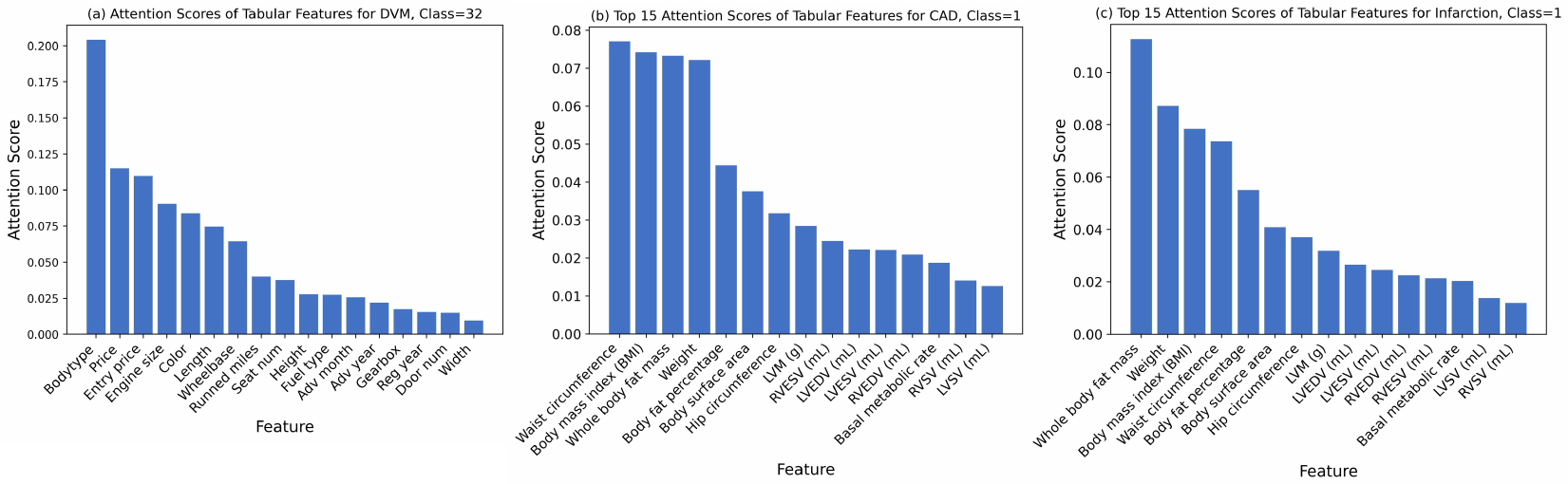}
  \caption{The [CLS] token's attention scores to tabular features for a particular class from the last layer of \modelname{}' tabular encoder. 
  }
  \label{fig:attention}
\end{figure}

\noindent\textbf{Visualization of \modelname{}'s Tabular Feature Attention}: In~\cref{fig:attention}, we visualize \modelname{}'s attention to different tabular features when predicting a specific class in downstream tasks. To achieve this, we average the self-attention map of samples belonging to the same class in test sets and present the attention scores of the [CLS] token. We observe that \modelname{} attends to not only image-related features, \eg, color in DVM, but also to features that are not directly visible in the image, \eg,  price in DVM. This showcases the importance of integrating multimodal data, as it can provide additional complementary information during downstream tasks, which can be difficult to obtain with image data alone. More visualization on cross-attention and case studies in Sec. C.5 of supp..

\section{Conclusion}
We have proposed \modelname{}, a novel tabular-image pre-training framework for multimodal representation learning. \modelname{} is a transformer-based multimodal network with a versatile tabular encoder and a multimodal interaction module, which are trained by a novel self-supervised pre-training strategy. In particular, \modelname{} accounts for tabular data missingness, which makes it applicable to real-world datasets. Experiments on natural and medical image datasets have showcased \modelname{}'s SOTA performance in various missing data scenarios and the efficacy of the proposed model components. The current work utilizes simulated missing data and 2D images. Future works will incorporate real-world incomplete data and extend to higher-dimensional images, \eg, 3D and temporal imaging data. Potential societal impact is discussed in the supplementary material.

\section*{Acknowledgements}
This research has been conducted using the UK Biobank Resource under Application Number 40616. The MR images presented in the figures are reproduced with the kind permission of UK Biobank ©. We also thank Paul Hager from the Lab for AI in Medicine at the Technical University of Munich for providing the pre-processing code for the UKBB dataset. DO'R is supported by the Medical Research Council (MC\_UP\_1605/13); National Institute for Health Research (NIHR) Imperial College Biomedical Research Centre; and the British Heart Foundation (RG/19/6/34387, RE/24/130023, CH/P/23/80008).

%
%
\bibliographystyle{splncs04}
\bibliography{egbib}

\newcounter{supsection}
\newcounter{supsubsection}[supsection]
\newcounter{supsubsubsection}[supsubsection]

\newcommand{\supsection}[1]{
    \stepcounter{supsection}
    \setcounter{supsubsection}{0}
    \section*{\Alph{supsection}. #1}
    \addcontentsline{toc}{section}{\Alph{supsection}. #1}
}

\newcommand{\supsubsection}[1]{
    \stepcounter{supsubsection}
    \setcounter{supsubsubsection}{0}
    \subsection*{\Alph{supsection}.\arabic{supsubsection}. #1}
    \addcontentsline{toc}{subsection}{\Alph{supsection}.\arabic{supsubsection}. #1}
}

\newcommand{\supsubsubsection}[1]{
    \stepcounter{supsubsubsection}
    \subsubsection*{\Alph{supsection}.\arabic{supsubsection}.\arabic{supsubsubsection}. #1}
    \addcontentsline{toc}{subsubsection}{\Alph{supsection}.\arabic{supsubsection}.\arabic{supsubsubsection}. #1}
}

\renewcommand{\thesection}{\Alph{supsection}}
\renewcommand{\thesubsection}{\Alph{supsection}.\arabic{supsubsection}}
\renewcommand{\thesubsubsection}{\Alph{supsection}.\arabic{supsubsection}.\arabic{supsubsubsection}}

\noindent\textbf{Broader Societal Impact}: Our proposed \modelname{}, which is optimized using statistical techniques, could potentially perpetuate biases and unfairness present in the training data. For instance, the image-tabular data in the UK Biobank database~\cite{bycroft2018uk} is mostly collected from white population and healthy subjects~\cite{bai2020population,sudlow2015uk}. Previous studies have found that deep learning models could potentially learn spurious correlations between cardiac diseases and population characteristics and may result in negative societal impacts when generalizing to other populations~\cite{ho2022ethnic,nayak2020understanding}. Therefore, further research and deployment based on this model should take into account these issues, along with potential solutions to address biases and unfairness.


\supsection{Detailed Data Description}

The UK Biobank (UKBB)~\cite{bycroft2018uk} is used for two cardiovascular disease classification (diagnosis) tasks: coronary artery disease (CAD) and myocardial infarction (Infarction). This dataset comprises 36,167 subjects. For each subject, we utilized mid-ventricle slices of its cardiac magnetic resonance (MR) images at three time phases, \ie, end-systolic (ES) frame, end-diastolic (ED) frame, and a time frame between ED and ES. In addition, we employed 75 tabular features, including 26 categorical features, \eg, alcohol drinker status, and 49 continuous features, \eg, average heart rate, and their detailed information can be found in~\cref{tab:UKBB}.

Moreover, we have used a natural image dataset, Data Visual Marketing (DVM)~\cite{huang2022dvm}, for a car model classification task with 283 classes. We employed 176,414 image-tabular samples from this dataset. As illustrated in~\cref{tab:DVM}, each sample has 17 tabular features in total, including 4 categorical and 13 continuous features. For both UKBB and DVM datasets, we pre-processed their tabular data before using them for our task as in~\cite{hager2023best}. Additionally, we converted categorical data into ordinal numbers and standardized continuous data using z-score normalization, with a mean value of 0 and standard derivation of 1.

\begin{table}[tb]
  \caption{75 tabular features (26 categorical and 49 continuous) are employed for CAD and Infarction tasks on UKBB. \textbf{Cat} denotes whether the feature is categorical, and \textbf{$N_{unq}$} represents the total number of unique values for each categorical feature. 
  }
  \label{tab:UKBB}
  \centering
  \resizebox{\textwidth}{!}{\begin{tabular}{p{78mm}P{6mm}P{7mm}|p{78mm}P{6mm}P{7mm}}
    \hline
    \textbf{Tabular Feature} & \textbf{Cat} & \textbf{$N_{unq}$} & \textbf{Tabular Feature} & \textbf{Cat} & \textbf{$N_{unq}$} \\
    \hline
    Alcohol drinker status & $\surd$ & 3 & LVCO (L/min) & $\times$ & - \\
    Alcohol intake frequency & $\surd$ & 6 & LVEDV (mL) & $\times$ & - \\
    Angina diagnosed by doctor & $\surd$ & 2 & LVEF (\%) & $\times$ & -  \\
    Augmentation index for PWA & $\times$ & - & LVESV (mL) & $\times$ & - \\
    Average heart rate & $\times$ & - & LVM (g) & $\times$ & - \\
    Basal metabolic rate & $\times$ & - & LVSV (mL) & $\times$ & - \\
    Blood pressure medication regularly taken & $\surd$ & 2 & Number of beats in waveform average for PWA & $\times$ & - \\
    Body fat percentage & $\times$ & - & \fontsize{7}{7}\selectfont Number of days/week of moderate physical activity 10+ minutes & $\surd$ & 8 \\
    Body mass index (BMI) & $\times$ & - & \fontsize{7}{7}\selectfont Number of days/week of vigorous physical activity 10+ minutes & $\surd$ & 8 \\
    Body surface area & $\times$ & - & Number of days/week walked 10+ minutes & $\surd$ & 8 \\
    Cardiac index during PWA & $\times$ & - & \fontsize{7}{7}\selectfont Oral contraceptive pill or minipill medication regularly taken & $\surd$ & 2  \\
    Cardiac index & $\times$ & - & Overall health rating & $\surd$ & 4 \\
    Cardiac output during PWA & $\times$ & - & P duration & $\times$ & - \\
    Cardiac output & $\times$ & - & Past tobacco smoking & $\surd$ & 4 \\
    Central augmentation pressure during PWA & $\times$ & - & Peripheral pulse pressure during PWA & $\times$ & - \\
    Central pulse pressure during PWA & $\times$ & - & Pulse rate & $\times$ & - \\
    Central systolic blood pressure during PWA & $\times$ & - & Pulse wave Arterial Stiffness index & $\times$ & - \\
    Cholesterol lowering medication regularly taken & $\surd$ & 2 & QRS duration & $\times$ & - \\
    Current tobacco smoking & $\surd$ & 3 & RVEDV (mL) & $\times$ & - \\
    Diabetes diagnosis & $\surd$ & 2 & RVEF (\%) & $\times$ & - \\
    Diastolic blood pressure & $\times$ & - & RVESV (mL) & $\times$ & - \\
    Diastolic brachial blood pressure during PWA & $\times$ & - & RVSV (mL) & $\times$ & - \\
    Duration of moderate activity & $\times$ & - & Sex & $\surd$ & 2 \\
    Duration of strenuous sports & $\surd$ & 8 & Shortness of breath walking on level ground & $\surd$ & 2 \\
    Duration of vigorous activity & $\times$ & - & Sleep duration & $\times$ & - \\
    Duration of walks & $\times$ & - & Sleeplessness / insomnia & $\surd$ & 3 \\
    End systolic pressure during PWA & $\times$ & - & Smoking status & $\surd$ & 3 \\
    End systolic pressure index during PWA & $\times$ & - & Stroke diagnosed by doctor &  $\surd$ & 2 \\
    Ever smoked & $\surd$ & 8 & Stroke volume during PWA & $\times$ & - \\
    Exposure to tobacco smoke at home & $\times$ & - & Systolic blood pressure &  $\times$ & -\\
    Exposure to tobacco smoke outside home & $\times$ & - & Systolic brachial blood pressure during PWA & $\times$ & - \\
    Falls in the last year  & $\surd$ & 3 & Total peripheral resistance during PWA & $\times$ & - \\
    Heart rate during PWA & $\times$ & - & Usual walking pace & $\times$ & - \\
    High blood pressure diagnosed by doctor & $\surd$ & 2 & Ventricular rate & $\times$ & - \\
    Hip circumference & $\times$ & - & Waist circumference & $\times$ & - \\
    \fontsize{8}{8}\selectfont Hormone replacement therapy medication regularly taken & $\surd$ & 2 & Weight & $\times$ & - \\
    Insulin medication regularly taken & $\surd$ & 2 & Whole body fat mass & $\times$ & - \\
    Long-standing illness, disability or infirmity & $\surd$ & 2 & & & \\
    \hline
  \end{tabular}}
\end{table}

\begin{table}[tb]
  \caption{17 tabular features (4 categorical and 13 continuous) are used for the DVM car model classification task. \textbf{Cat} denotes whether the feature is categorical,  and \textbf{$N_{unq}$} represents the total number of unique values for each categorical feature. 
  }
  \label{tab:DVM}
  \centering
  \resizebox{\textwidth}{!}{\begin{tabular}{p{78mm}P{6mm}P{7mm}|p{78mm}P{6mm}P{7mm}}
    \hline
    \textbf{Tabular Feature} & \textbf{Cat} & \textbf{$N_{unq}$} & \textbf{Tabular Feature} & \textbf{Cat} & \textbf{$N_{unq}$} \\
    \hline
    Advertisement month (Adv\_month) & $\times$ & - & Height & $\times$ & - \\
    Advertisement year (Adv\_year) & $\times$ & - & Length & $\times$ & - \\
    Bodytype & $\surd$ & 13 & Price & $\times$ & - \\
    Color & $\surd$ & 22 & Registration year (Reg\_year) & $\times$ & - \\
    Number of doors (Door\_num) & $\times$ & - & Miles runned (Runned\_Miles) & $\times$ & - \\
    Engine size (Engine\_size) & $\times$ & - & Number of seats (Seat\_num) & $\times$ & - \\
    Entry prize (Entry\_prize) & $\times$ & - & Wheelbase & $\times$ & - \\
    Fuel type (Fuel\_type) & $\surd$ & 12 & Width & $\times$ & - \\
    Gearbox & $\surd$ & 3 & & & \\
    \hline
  \end{tabular}}
\end{table}

\supsection{Implementation Details}
\supsubsection{Pre-training}
We followed the same image and tabular data augmentation techniques during pre-training as in~\cite{hager2023best}. Specifically, we augmented images through random scaling, rotation, shifting, flipping, Gaussian noise, as well as brightness, saturation, and contrastive changes. After that, all images are resized to $128 \times 128$. To speed up the image augmentation process, we used the Albumentations python library~\cite{buslaev2020albumentations}. For tabular data augmentation of ITC and ITM, we randomly selected 30\% of tabular features in each subject and replaced their values with column-wise randomly selected values. Notice that tabular pre-training algorithms (SCARF~\cite{bahri2022scarf}, VIME~\cite{yoon2020vime}, and SAINT~\cite{somepalli2021saint}) have their own tabular augmentations.

The hyper-parameters and training configurations for the self-supervised learning (SSL) image pre-training approaches (SimCLR~\cite{chen2020simple}, BYOL~\cite{grill2020bootstrap}, SimSiam~\cite{chen2021exploring}, BarlowTwins~\cite{zbontar2021barlow}) and SSL multimodal pre-training method (MMCL~\cite{hager2023best}) are the same as those used in~\cite{hager2023best}, which were found using hyper-parameter search. We utilized optimal hyper-parameters to pre-train SSL tabular pre-training models and our proposed \modelname{}. Specifically, for each model, we selected the best learning rate from a set of values of $\{3 \times 10^{-3}, 3 \times 10^{-4}, 3 \times 10^{-5}\}$ and the best weight decay from $\{1 \times 10^{-4},1.5 \times 10^{-6}\}$, based on its performance on the validation set. All models are deployed on 4 A6000 GPUs and pre-trained for 500 epochs using the Adam optimizer~\cite{kingma2014adam}. The learning rate is warmed up linearly for 10 epochs and decayed following a cosine annealing scheduler. The implementation details of \modelname{} and SSL tabular pre-training methods are discussed below.

\noindent\textbf{The Proposed \modelname{}:} It utilizes a learning rate of $3 \times 10^{-4}$ and a weight decay of $1.5 \times 10^{-6}$ for DVM pre-training and a learning rate of $3 \times 10^{-4}$ and a weight decay of $1 \times 10^{-4}$ for UKBB cardiac pre-training. 


\noindent\textbf{SCARF~\cite{bahri2022scarf}:} It applies contrastive learning to original tabular data and an augmented view by corrupting a random subset of features. Based on the validation performance, the corruption ratio and the temperature parameter are set to 0.3 and 0.1, respectively. The hidden dimension of SCARF's multi-layer perceptron (MLP) is 512. We utilized a learning rate of $3 \times 10^{-4}$ and a weight decay of $1.5 \times 10^{-6}$ for DVM pre-training and a learning rate of $3 \times 10^{-3}$ and a weight decay of $1 \times 10^{-4}$ for UKBB cardiac pre-training. 

\noindent\textbf{VIME~\cite{yoon2020vime}:} It predicts the corrupted positions in tabular data and reconstructs their values. Based on the validation performance, the corruption ratio is set to 0.3. The reconstruction loss adjustment parameter $\alpha$ is 2.0 as in~\cite{yoon2020vime}, and the hidden dimension of VIME's MLP is 512. For pre-training, we utilized a learning rate of $3 \times 10^{-4}$ and a weight decay of $1.5 \times 10^{-6}$ for DVM and a learning rate of $3 \times 10^{-3}$ and a weight decay of $1 \times 10^{-4}$ for UKBB cardiac dataset. 

\noindent\textbf{SAINT~\cite{somepalli2021saint}:} It produces an augmented tabular view through CutMix~\cite{yun2019cutmix} and mixup~\cite{zhang2018mixup} and then operates contrastive learning and denoising pre-training. Additionally, SCARF proposed a new transformer-based tabular architecture that performs attention across rows and columns. We utilized a learning rate of $3 \times 10^{-5}$ and a weight decay of $1.5 \times 10^{-6}$ for DVM pre-training and a learning rate of $3 \times 10^{-5}$ and a weight decay of $1 \times 10^{-4}$ for UKBB cardiac pre-training. 

\supsubsection{Fine-tuning}
For either fully fine-tuning or linear probing settings of each pre-trained model, we chose the best learning rate from a set of values of $\{3 \times 10^{-2}, 1 \times 10^{-2}, 3 \times 10^{-3}, 1 \times 10^{-3}, 3 \times 10^{-4}, 1 \times 10^{-4}, 3 \times 10^{-5}, 1 \times 10^{-5}\}$ depending on its validation performance. \cref{tab:learning_rate}(b) demonstrates the learning rate used for each model. We utilized an Adam optimizer without weight decay and a batch size of 512. To alleviate over-fitting, an early stopping strategy in Pytorch Lightning has been adopted, with a minimal delta (divergence threshold) of 0.0002, a maximal number of epochs of 500, and a patience (stopping threshold) of 10 epochs. 


\begin{table}[tb]
  \caption{Learning rates of DVM, CAD, and Infarction tasks for different models during supervised training or fine-tuning. \faSnowflake[regular] means linear probing, and \faFire* represents fully fine-tuning.
  }
  \label{tab:learning_rate}
  \centering
  \resizebox{\textwidth}{!}{\begin{tabular}{p{35mm}|P{17mm}P{17mm}|P{17mm}P{17mm}|P{17mm}P{17mm}}
    \hline
    Model & \multicolumn{2}{c|}{DVM Accuracy (\%) $\uparrow$} & \multicolumn{2}{c|}{CAD AUC (\%) $\uparrow$} & \multicolumn{2}{c}{Infarction AUC (\%) $\uparrow$} \\
    \cline{2-7}
    ~ & \small\faSnowflake[regular] & \faFire* &  \faSnowflake[regular] & \faFire* &  \faSnowflake[regular] & \faFire* \\
    \hline
    \multicolumn{7}{c}{(a) Supervised Methods} \\ 
    \hline
    ResNet-50~\cite{he2016deep}  & \multicolumn{2}{c|}{$3 \times 10^{-4}$} & \multicolumn{2}{c|}{$1 \times 10^{-3}$} & \multicolumn{2}{c}{$1 \times 10^{-3}$} \\
    Concat Fuse 
    (CF)~\cite{spasov2019parameter} & \multicolumn{2}{c|}{$3 \times 10^{-4}$} & \multicolumn{2}{c|}{$3 \times 10^{-3}$} & \multicolumn{2}{c}{$3 \times 10^{-3}$} \\
    Max Fuse (MF)~\cite{vale2021long}  & \multicolumn{2}{c|}{$3 \times 10^{-4}$} & \multicolumn{2}{c|}{$3 \times 10^{-3}$} & \multicolumn{2}{c}{$3 \times 10^{-3}$} \\
    Interact Fuse (IF)~\cite{duanmu2020prediction}  & \multicolumn{2}{c|}{$3 \times 10^{-4}$} & \multicolumn{2}{c|}{$3 \times 10^{-3}$} & \multicolumn{2}{c}{$3 \times 10^{-3}$} \\
    DAFT~\cite{wolf2022daft} & \multicolumn{2}{c|}{$3 \times 10^{-4}$} & \multicolumn{2}{c|}{$3 \times 10^{-3}$} & \multicolumn{2}{c}{$3 \times 10^{-3}$} \\
    \hline
    \multicolumn{7}{c}{(b) SSL Pre-training Methods} \\ 
    \hline
    SimCLR~\cite{chen2020simple} & $1 \times 10^{-3}$ & $1 \times 10^{-4}$ & $1 \times 10^{-3}$ & $1 \times 10^{-3}$ & $1 \times 10^{-3}$ & $1 \times 10^{-3}$ \\
    BYOL~\cite{grill2020bootstrap} & $1 \times 10^{-3}$ & $1 \times 10^{-4}$ & $1 \times 10^{-3}$ & $1 \times 10^{-4}$ & $1 \times 10^{-3}$ & $1 \times 10^{-4}$\\
    SimSiam~\cite{chen2021exploring} & $1 \times 10^{-3}$ & $1 \times 10^{-5}$ & $1 \times 10^{-3}$ & $1 \times 10^{-4}$ & $1 \times 10^{-3}$ & $1 \times 10^{-4}$ \\
    BarlowTwins~\cite{zbontar2021barlow} & $1 \times 10^{-3}$ & $1 \times 10^{-3}$ & $1 \times 10^{-3}$ & $1 \times 10^{-3}$ & $1 \times 10^{-3}$ & $1 \times 10^{-3}$ \\
    SCARF~\cite{bahri2022scarf} & $1 \times 10^{-4}$ & $1 \times 10^{-4}$ & $1 \times 10^{-3}$ & $1 \times 10^{-3}$ & $1 \times 10^{-3}$ & $1 \times 10^{-3}$ \\
    VIME~\cite{yoon2020vime} & $1 \times 10^{-4}$ & $1 \times 10^{-4}$ & $1 \times 10^{-3}$ & $1 \times 10^{-3}$ & $1 \times 10^{-3}$ & $1 \times 10^{-3}$ \\
    SAINT~\cite{somepalli2021saint} & $1 \times 10^{-4}$ & $1 \times 10^{-5}$ & $1 \times 10^{-3}$ & $1 \times 10^{-5}$ & $1 \times 10^{-3}$ & $1 \times 10^{-5}$ \\
    MMCL~\cite{hager2023best} & $1 \times 10^{-3}$ & $1 \times 10^{-3}$ & $1 \times 10^{-3}$ & $1 \times 10^{-3}$ & $1 \times 10^{-3}$ & $1 \times 10^{-3}$ \\
    \modelname{} (proposed) & $1 \times 10^{-4}$ & $1 \times 10^{-4}$ & $1 \times 10^{-3}$ & $1 \times 10^{-4}$ & $1 \times 10^{-3}$ & $1 \times 10^{-4}$ \\
    \hline
  \end{tabular}}
\end{table}

\supsubsection{Supervised Training}
We reproduced 1 supervised image approach, ResNet-50, and 4 supervised multimodal algorithms: concatenation fusion (CF)~\cite{spasov2019parameter}, maximum fusion (MF)~\cite{vale2021long}, interactive fusion through channel-wise multiplication (IF)~\cite{duanmu2020prediction}, and dynamic affine transform (DAFT)~\cite{wolf2022daft}. These multimodal techniques leverage a ResNet-50 as their image encoder for fair comparison. To adapt CF and MF to our task, we used a 2-layer MLP with a hidden dimension of 512 and an output dimension of 2048 as their tabular encoder. For IF, its tabular encoder is a 4-layer MLP, with hidden dimensions of [64, 256, 512, 1024] and an output dimension of 2048. We undertook the same training strategy and learning rate sweep as the fine-tuning process, \ie, an early stopping strategy with a maximum of 500 epochs. We ensure that all supervised models were converged after training. \cref{tab:learning_rate}(a) demonstrates the learning rate used for each model. 

\supsection{Additional Experiment}

\begin{figure}[tb]
  \centering
  \includegraphics[width=1\linewidth]{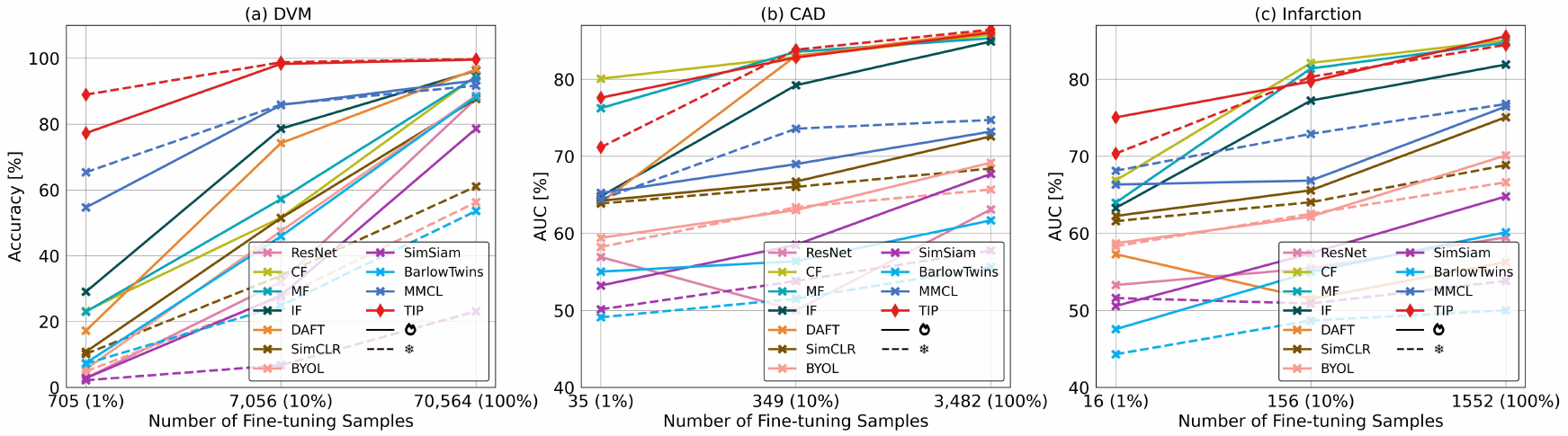}
  \caption{
Overall result comparison with supervised/SSL image/multimodal approaches on various number of fine-tuning samples. \faFire* denotes fully fine-tuning, and \faSnowflake[regular] means linear probing. In addition to the results shown in Fig. 3 of the manuscript, we have included the results of BYOL, SimSiam, and BarlowTwins.
  }
  \label{fig:sup_low}
\end{figure}

\supsubsection{Robustness to Low-data Regimes (Complete Results)}
As mentioned in Sec. 4.1 of the manuscript, we propose to assess the performance of TIP and other SOTA methods on low-data regimes (10\% and 1\% of the original dataset size). Fig. 3 in the manuscript displays only SimCLR's results for SSL image approaches since it showed the best performance among them. We present the complete results of all models in~\cref{fig:sup_low}.

\supsubsection{Ablation Study on The Proposed SSL Strategy}
We conduct an ablation study to analyze the impact of each SSL pre-training task: image-tabular contrastive learning (ITC), image-tabular matching (ITM), and masked tabular reconstruction (MTR). \cref{tab:sup_ablation} demonstrates the results on complete downstream task data. We can obtain the following observations: (1) Compared with supervised \modelname{} w/o any pre-training tasks, adding pre-training tasks improves the model performance, indicating the usefulness of our pre-training strategy. (2) In the linear probing setting, compared with integrated \modelname{}, removing any of our pre-training tasks significantly decreases the model performance, \eg, \modelname{} w/o ITC decreases AUC by 14.08\% on Infarction, \modelname{} w/o ITM decreases AUC by 1.61\% on CAD, and \modelname{} w/o MTR decreases AUC by 2\% on CAD. This indicates that our three pre-training tasks enable the model to learn transferable features and efficiently produce promising results with a few tunable parameters. The competitive results of \modelname{} and \modelname{} w/o ITM or ITC in fully fine-tuning can be attributed to the relatively small pre-training datasets and also the fact that tuning all parameters can moderately alleviate the reliance on integrating all three pre-training tasks.

\begin{figure}[tb]
  \centering
  \includegraphics[width=0.9\linewidth]{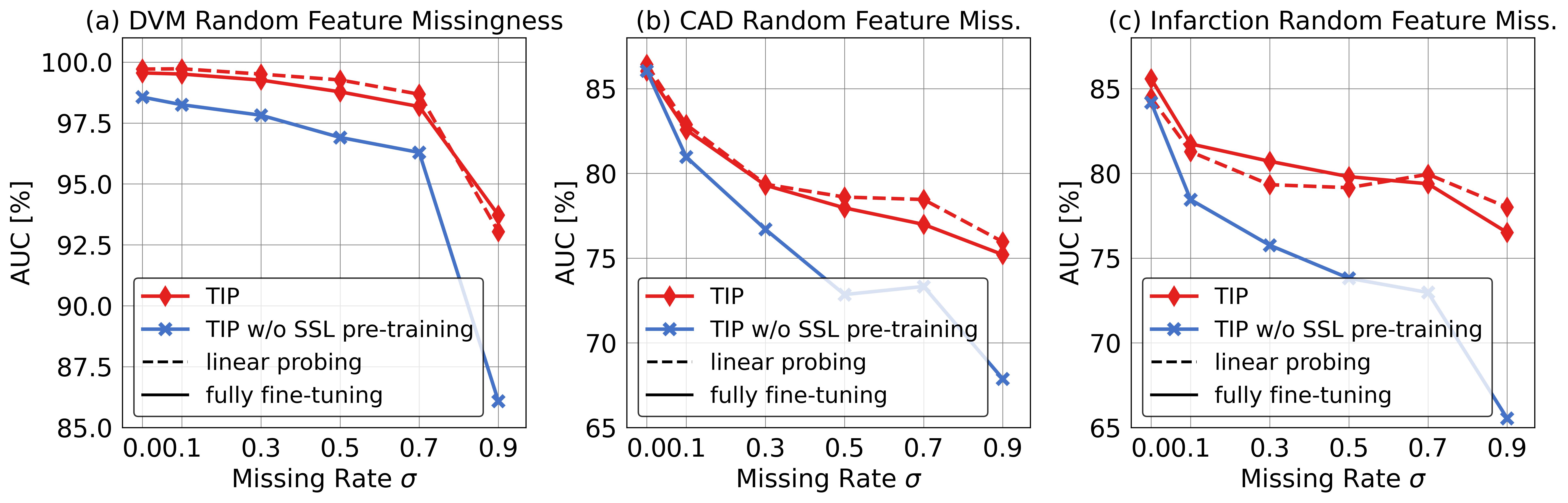}
  \caption{Results of DVM, CAD, and Infarction tasks comparing \modelname{} with or without the proposed SSL pre-training in the random feature missingness (RFM) scenario with different missing rates. In addition to the DVM's and CAD's results shown in Fig. 5 of the manuscript, we have included the results on Infarction.
  }
  \label{fig:sup_missingness_noSSL}
\end{figure}

Moreover, as mentioned in Sec. 4.3 of the manuscript, we study the performance of \modelname{} with and without our SSL pre-training strategy when encountering incomplete downstream task data. Fig. 5 in the manuscript only illustrates the results on DVM and CAD due to page limitations. We present the complete results of DVM, CAD, and Infarction in~\cref{fig:sup_missingness_noSSL}. We observe that our pre-training task enhances the model robustness to missing data across various missing rates on DVM, CAD, and Infarction tasks.

\begin{table}[tb]
  \caption{Ablation study of \modelname{}'s SSL pre-training tasks on complete data. \faSnowflake[regular] means linear probing, and \faFire* represents fully fine-tuning. \modelname{} w/o pre-training (1st row) is trained in a supervised manner, \ie, all of its parameters are trainable in both \faSnowflake[regular] and \faFire* columns.
  }
  \label{tab:sup_ablation}
  \centering
  \resizebox{\textwidth}{!}{\begin{tabular}{P{10mm}|P{10mm}|P{10mm}|P{16mm}P{16mm}|P{16mm}P{16mm}|P{16mm}P{16mm}}
    \hline
    ITC & ITM & MTR & \multicolumn{2}{c|}{DVM Accuracy (\%) $\uparrow$} & \multicolumn{2}{c|}{CAD AUC (\%) $\uparrow$} & \multicolumn{2}{c}{Infarction AUC (\%) $\uparrow$} \\
    \cline{4-9}
    ~ & ~ & ~ & \small\faSnowflake[regular] & \faFire* &  \faSnowflake[regular] & \faFire* &  \faSnowflake[regular] & \faFire* \\
    \hline
    ~ & ~ & ~ & 98.57 & 98.57 & 86.04 & 86.04 & 84.19 & 84.19\\
    \hline
    ~ & $\surd$ & $\surd$ & 98.84 & 99.14 & 76.51 & \textbf{86.89} & 70.38 & 85.72 \\
    $\surd$ & ~ & $\surd$ & 99.71 & 99.53 & 84.82 & 86.22 & 83.71 & \textbf{85.89} \\
    $\surd$ & $\surd$ & ~ & 99.70 & 99.56 & 84.43 & 86.11 & 82.91 & 85.78 \\
    \rowcolor{gray!25}
    $\surd$ & $\surd$ & $\surd$ & \textbf{99.72} & \textbf{99.56} & \textbf{86.43} & 86.03 & \textbf{84.46} & 85.58 \\
  \hline
  \end{tabular}}
\end{table}

\supsubsection{Effect of TIP's Tabular Encoder}
We examine the contributions of our proposed transformer-based tabular encoder to the performance increase compared to supervised multimodal methods. Specifically, we replaced the MLP-based tabular encoder in the supervised multimodal methods (CF and MF) with the tabular encoder from \modelname{} and conducted incomplete data experiments on the DVM classification task. As shown in~\cref{fig:DVM_cross_attention}, \modelname{}'s tabular encoder can improve the performance of supervised multimodal methods. However, these methods still lag behind \modelname{}, especially in a high missing rate condition, demonstrating the efficacy of other components of \modelname{}.

\begin{figure}[tb]
  \centering
  \includegraphics[width=0.8\linewidth]{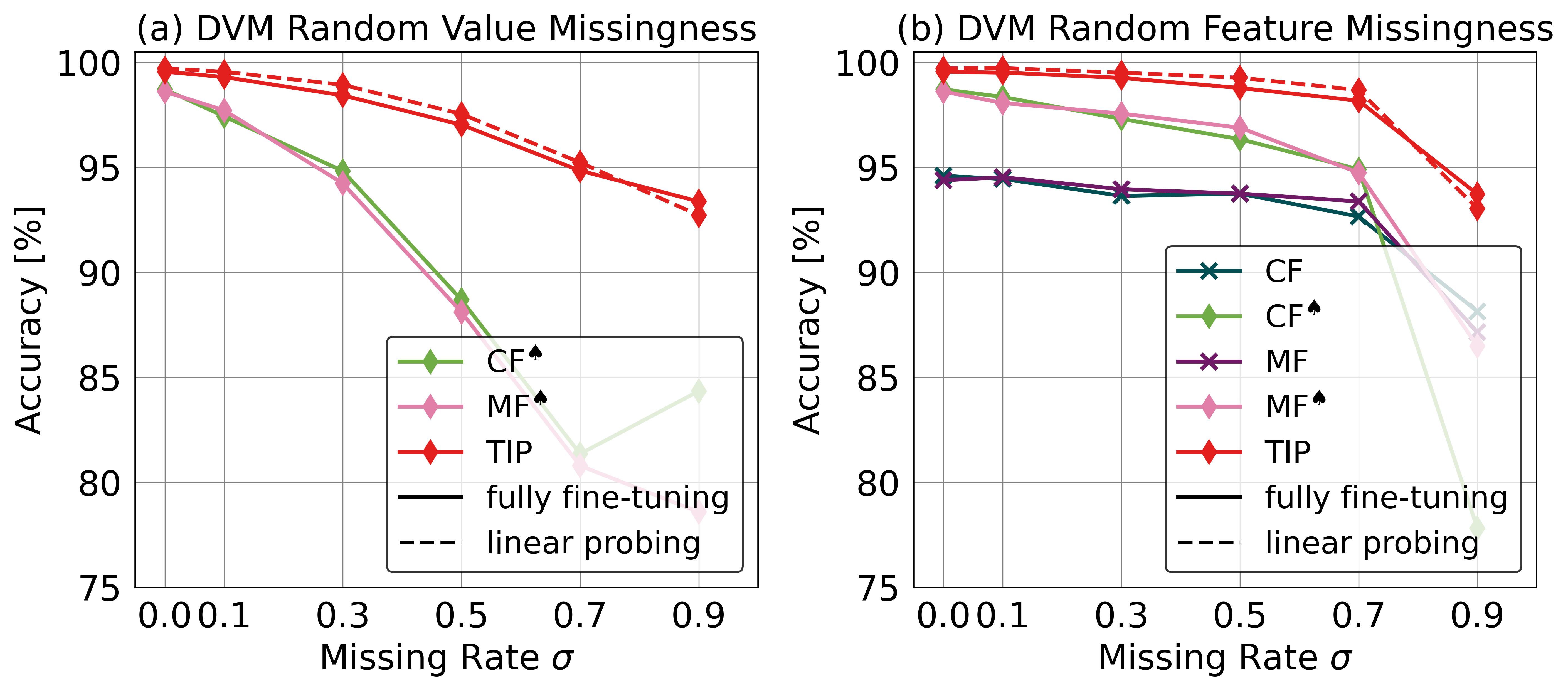}
  \caption{Results comparing supervised multimodal methods and TIP on the DVM random value missingness (RVM) and random feature missingness (RFM) scenarios. $^{\spadesuit}$ means using \modelname{}'s tabular encoder.
  }
  \label{fig:missingness_tabular}
\end{figure}

\begin{figure}[tb]
  \centering
  \includegraphics[width=1\linewidth]{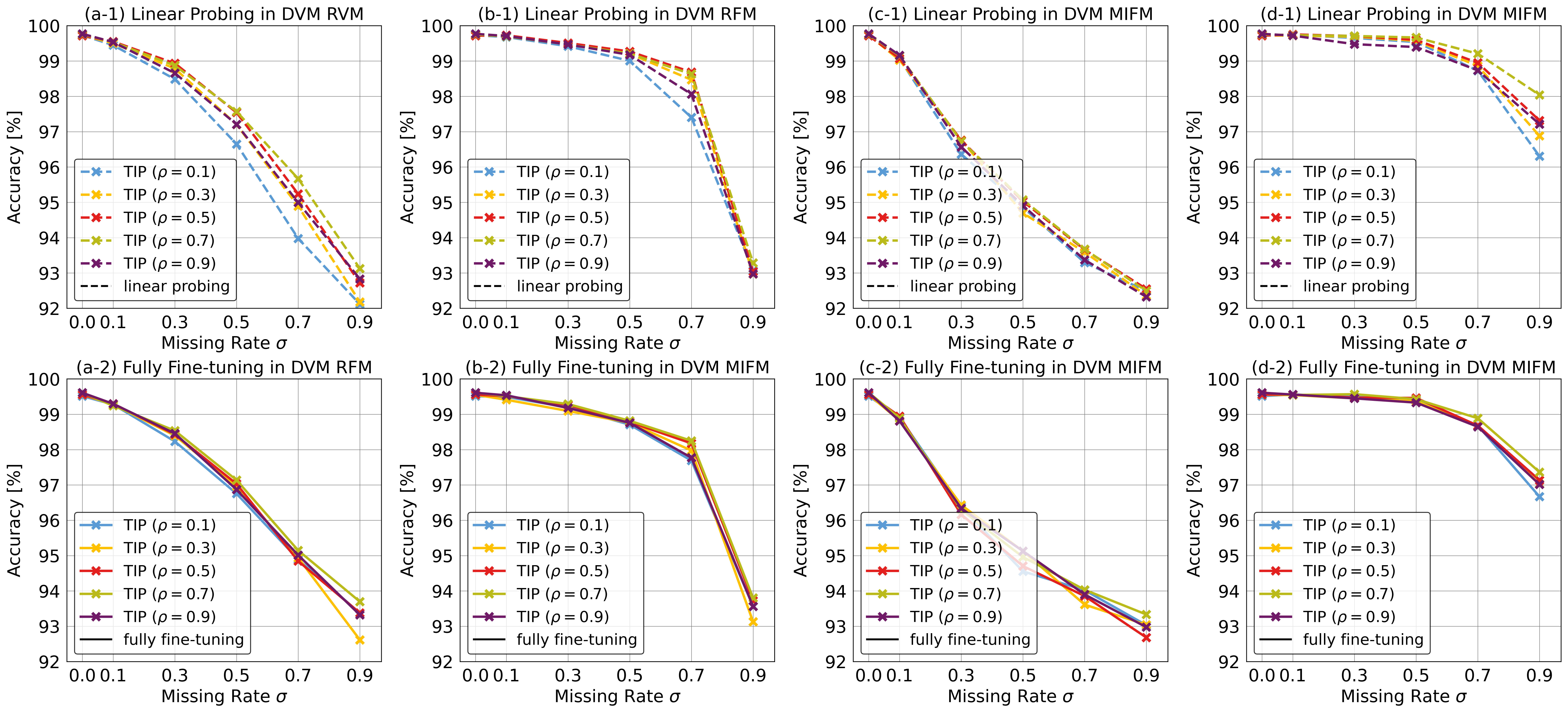}
  \caption{Results of \modelname{} with different masking ratios $\rho$ on 4 DVM's missing data scenarios: (a) random value missingness (RVM), (b) random feature missingness (RFM), (c) most important feature missingness (MIFM), and (d) least important feature missingness (LIFM). We evaluated linear probing (1st row) and fully fine-tuning (2nd row). Notice that $\rho$ is the masking ratio of the MTR pre-training task, while $\sigma$ is the missing rate of missing data scenarios.
  }
  \label{fig:missingness_mask}
\end{figure}

\supsubsection{Sensitivity of Masking Ratio}
As mentioned in Sec. 4.3 of the manuscript, we evaluated the effect of different masking ratios of the MTR pre-training task. In addition to Tab. 4 in the manuscript, we present the results of missing value reconstruction on two datasets in~\cref{tab:sup_imputation} and conducted experiments on the DVM classification task using diverse masking ratios (\cref{fig:missingness_mask}). As shown in~\cref{tab:imputation}, $\rho \in {(0.5, 0.7)}$ achieve the best reconstruction performance, whereas too high (0.9) or too low (0.1) masking ratios adversely affect model learning. In addition, the higher RMSE in UKBB than that in DVM indicates that there could be some outliers in UKBB. However, compared with SOTA data imputation methods in Tab. 2 of the manuscript, our \modelname{} still achieves the best performance. 

As displyed in~\cref{fig:missingness_mask}, \modelname{} has fairly consistent results in data missing scenarios across masking ratios, and moderate ratios (0.3, 0.5, 0.7) are better than extreme ones (0.1, 0.9). The sensitivity of $\rho$ in fully fine-tuning is smaller than that in linear probing. This may be because tuning all the parameters mitigates the reliance on optimal masking ratios.

\begin{table}[tb]
  \caption{\modelname{}’s RMSE results on the DVM and UKBB test sets for reconstruction of missing continuous features. $\sigma$ denotes data missing rate in fine-tuning and inference, and $\rho$ means masking ratio of the MTR pre-training task.}
  \label{tab:sup_imputation}
  \centering
  \resizebox{\textwidth}{!}{\begin{tabular}{p{27mm}|P{17mm}P{17mm}P{17mm}|P{17mm}P{17mm}P{17mm}}
  \hline
  Model & \multicolumn{3}{c|}{DVM RMSE $\downarrow$} & \multicolumn{3}{c}{UKBB RMSE $\downarrow$} \\
  \cline{1-7}
  Missing rate $\sigma$ & 0.3 & 0.5 & 0.7 & 0.3 & 0.5 & 0.7 \\
  \hline
  $\rho=0.1$ & 0.5349 & 0.6752 & 0.7871 & 0.6245 & 0.6903 & 0.7851 \\
  $\rho=0.3$ & 0.4110 & 0.5128 & 0.5924 & 0.6044 & 0.6538 & 0.7469 \\
  $\rho=0.5$ & \textbf{0.3899} & 0.4651 & 0.5055 & 0.6039 & 0.6460 & 0.7106 \\
  $\rho=0.7$ & 0.3986 & \textbf{0.4612} & \textbf{0.4733} & \textbf{0.5963} & \textbf{0.6171} & \textbf{0.6654} \\
  $\rho=0.9$ & 0.4279 & 0.4800 & 0.4816 & 0.6542  & 0.6696 & 0.6791  \\
  \hline
  \end{tabular}}
\end{table}

\begin{figure}[tb]
  \centering
  \includegraphics[width=1\linewidth]{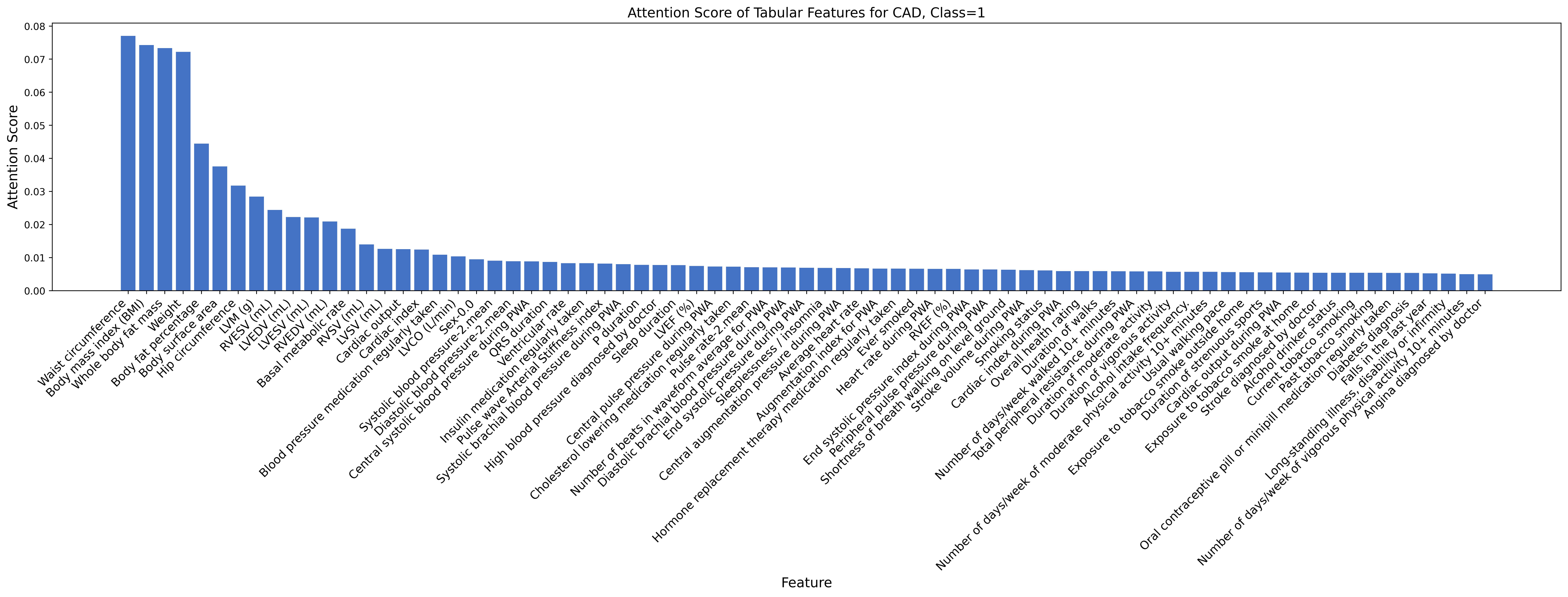}
  \caption{The [CLS] token's attention scores to tabular features for the True class in the CAD task from the last layer of \modelname{}' tabular encoder. 
  }
  \label{fig:attention_CAD}
\end{figure}

\begin{figure}[tb]
  \centering
  \includegraphics[width=1\linewidth]{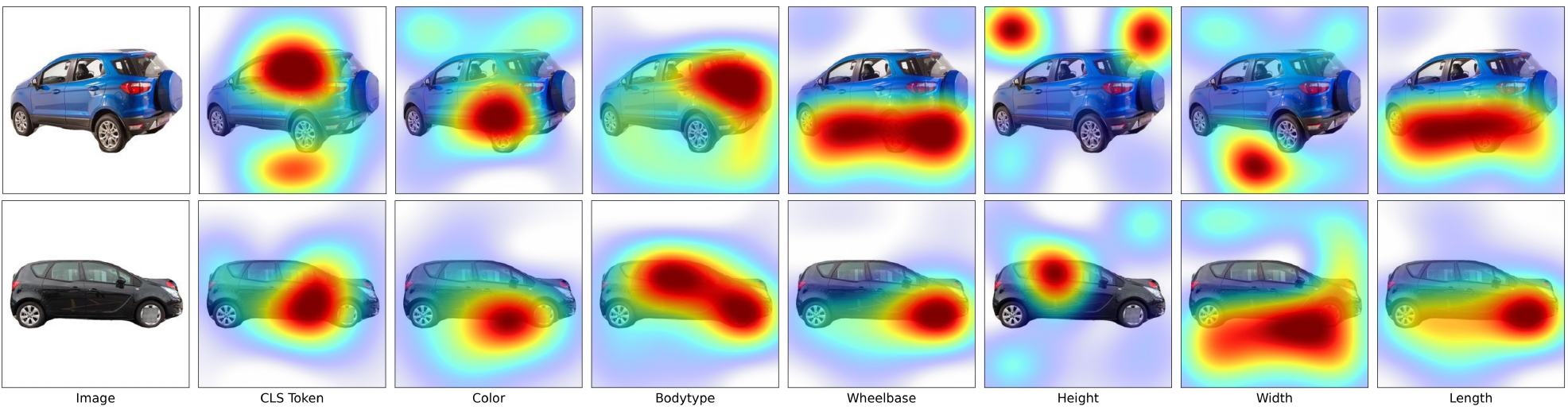}
  \caption{
Grad-CAM visualization on the cross-attention map in the 2nd layer of \modelname{}'s multimodal interaction module.
  }
  \label{fig:DVM_cross_attention}
\end{figure}

\supsubsection{Visualization}
As mentioned in Sec. 4.3 of the manuscript, we illustrate \modelname{}'s attention to tabular features when predicting a specific class in downstream tasks. \cref{fig:attention_CAD} shows complete attention scores to all tabular features in downstream CAD task. Our observations are as follows: (1) \modelname{} distinguishes important imaging phenotypes, \eg, the model attends more to left ventricle myocardial mass (LVM) and left ventricle end-diastolic volume (LVEDV) than left ventricle ejection fraction (LVEF), which is consistent with previous cardiac disease studies~\cite{bai2020population}. (2) \modelname{} attends to critical non-imaging risk factors, \eg, obesity-related features such as waist circumference and whole body fact mass~\cite{wilkins2016discordance,sniderman2019apolipoprotein}. (3) \modelname{} focuses more on physical measurements, \eg, weight, body fat percentage, and blood pressure. These measurements have demonstrated high correlations with the left ventricular function, which plays an important role in CAD diagnosis~\cite{bai2020population}. 

Furthermore, we computed Grad-CAM~\cite{li2021align,selvaraju2017grad} on the cross-attention maps in the 2nd layer of \modelname{}'s multimodal interaction module and generated per-token visualization. As displayed in~\cref{fig:DVM_cross_attention}, \modelname{} does not only identify the classification object, but also captures inter-modality relations, \eg, the `Bodytype' token attends to the entire car, whereas the `Wheelbase' token mainly focuses on the wheels. This showcases the effectiveness of our multimodal interaction module. 

\begin{figure}[tb]
  \centering
  \includegraphics[width=1\linewidth]{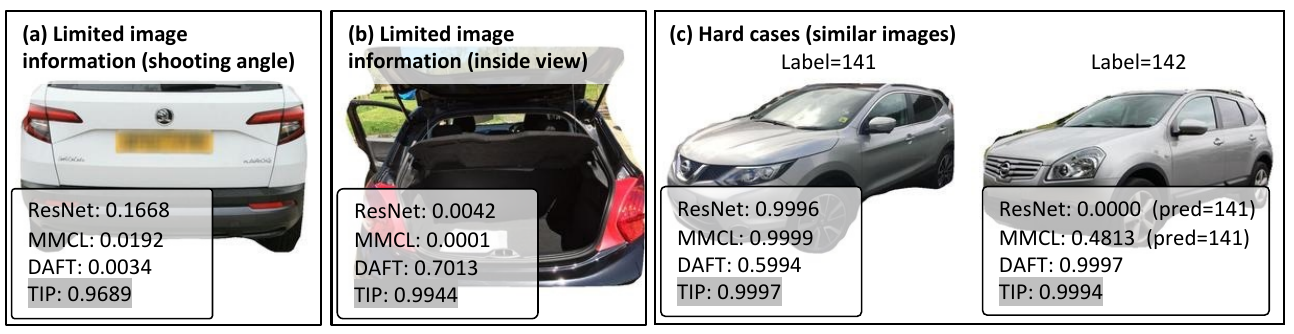}
  \caption{DVM car visualization of samples and ground-truth class's predictions of \modelname{} and other methods.
  }
  \label{fig:case_study}
\end{figure}

We also visualize some challenging cases of the DVM classification task where \modelname{} still outperforms supervised/SSL image/multimodal algorithms in~\cref{fig:case_study}. The results demonstrate that a single image modality may not provide sufficient information for decision-making, whereas \modelname{} can effectively integrate multimodal information to enhance model performance.

\end{document}